\documentclass[11pt]{article}

\usepackage[final]{acl}

\usepackage{times}
\usepackage{latexsym}
\usepackage{enumitem}
\usepackage{titlesec}
\usepackage[T1]{fontenc}
\usepackage[utf8]{inputenc}

\usepackage{microtype}

\usepackage{inconsolata}

\usepackage{graphicx}

\usepackage{amsmath}
\usepackage{booktabs}
\usepackage{subcaption}
\usepackage{algorithm}
\usepackage{algpseudocode}

\usepackage{enumitem}
\setlist{nosep}
\usepackage[most]{tcolorbox} 
\usepackage{array}
\usepackage{placeins}
\usepackage{stfloats}
\usepackage{needspace}
\usepackage{float}
\usepackage{amsfonts}
\usepackage{setspace}

\title{Beyond Fixed Psychological Personas: State Beats Trait, but Language Models are State-Blind}
\author{
  \textbf{Tamunotonye Harry\textsuperscript{1}}, 
  \textbf{Ivoline Ngong\textsuperscript{1}}, 
  \textbf{Chima Nweke\textsuperscript{2}}, 
  \textbf{Yuanyuan Feng\textsuperscript{1}}, 
  \textbf{Joseph Near\textsuperscript{1}} 
\\
  \textsuperscript{1}University of Vermont, Burlington, VT, USA \\
  \textsuperscript{2}Independent Researcher \\
  \small{
    \textbf{Correspondence:} \href{mailto:tamunotonye.harry@uvm.edu}{tamunotonye.harry@uvm.edu}
    }
}
\begin{document}
\maketitle
\begin{abstract}
% The same user expresses different psychological states across contexts—anxious when seeking medical advice, confident when discussing hobbies. Yet existing persona datasets treat psychology as static: they capture only stable traits (typically Big Five personality), from single contexts, assuming users have fixed profiles. We introduce Chameleon, a dataset of 5,001 contextual psychological profiles from 1,667 Reddit users, each measured across multiple contexts. Unlike prior work, Chameleon spans 26 validated dimensions across four psychological frameworks—personality, values, motivation, and risk attitudes—enabling analysis of how the same individual varies across situations.
% Inspired by Latent State-Trait theory, we decompose variance and find that 74\% is within-person (state) while only 26\% is between-person (trait): context shapes expressed psychology 2–3x more than individual differences. Cross-method agreement and literature-driven hypothesis tests validate these patterns. Do AI systems appropriately handle this psychological variation? Two experiments suggest they get it backwards. In generation, LLMs produce highly similar responses regardless of stated user profile—they are state-blind when personalization would be appropriate. In evaluation, reward models assign systematically different scores to identical responses based on user profile—they are state-biased when invariance is required. AI systems ignore psychological context when they should adapt, and penalize it when they should not. We release Chameleon to support psychology-aware AI research.

User interactions with language models vary due to static properties of the user (trait) and the specific context of the interaction (state). However, existing persona datasets (like PersonaChat, PANDORA etc.) capture only trait, and ignore the impact of state. We introduce Chameleon, a dataset of 5,001 contextual psychological profiles from 1,667 Reddit users, each measured across multiple contexts. Using the Chameleon dataset, we present three key findings. First, inspired by Latent State-Trait theory, we decompose variance and find that 74\% is within-person(state) while only 26\% is between-person (trait). Second, we find that LLMs are state-blind: they focus on trait only, and produce similar responses regardless of state. Third, we find that reward models react to user state, but inconsistently: different models favor or penalize the same users in opposite directions. We release Chameleon to support research on affective computing, personalized dialogue, and RLHF alignment.

\end{abstract}

\section{Introduction}
\begin{figure}[t]
\centering
\includegraphics[width=\columnwidth]{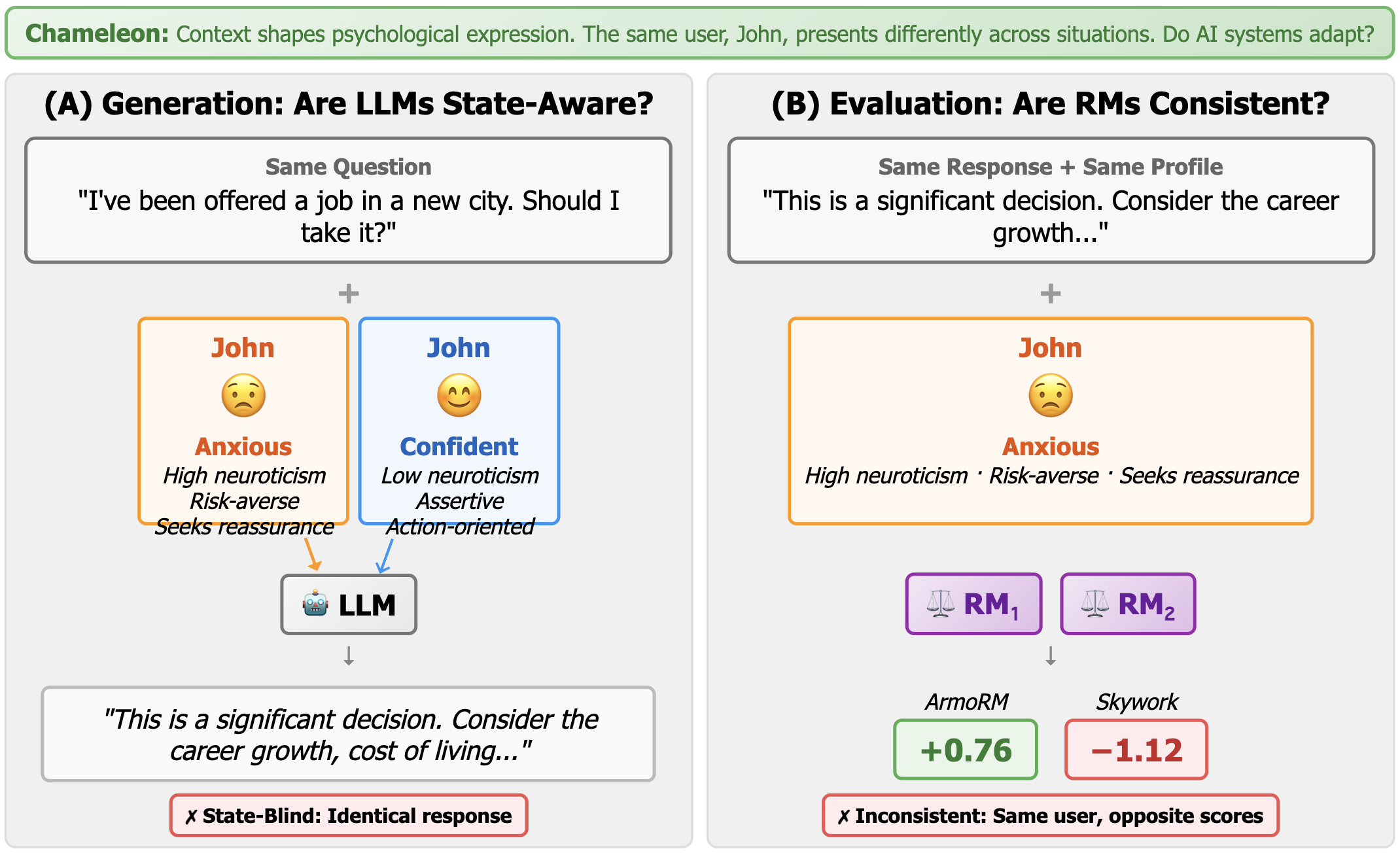}
\caption{\textbf{AI systems get psychological context backwards.} The same user (John) expresses different psychological states across contexts (74\% of variance is within-person). (A) \textbf{Generation:} LLMs produce nearly identical responses regardless of user profile. (B) \textbf{Evaluation:} Reward models score identical responses differently based on user profile, but disagree on direction. LLMs are state-blind; reward models are context-aware but inconsistent.}
\label{fig:problem_1}
\end{figure}

Decades of research in Latent State-Trait (LST) theory demonstrates that human behavior reflects both stable \textit{traits} (enduring characteristics like extraversion) and contextual \textit{states} (momentary expressions shaped by situation) \citep{steyer1999latent,fleeson2001toward}. Each interaction with an LLM is no different: users bring both enduring characteristics and momentary expressions shaped by their current situation. The same person expresses different psychological characteristics across different contexts, not because they changed, but because context shapes expression. 

Consider two emails from John, a student seeking guidance. The first reflects an \textbf{anxious \textit{state}}: \textit{"I've been stuck on this for hours and I'm starting to panic. I've reread everything and nothing clicks. I'm sorry to bother you, maybe I'm just not cut out for this."} The second reflects a \textbf{confident \textit{state}}: \textit{"I've narrowed it down to two directions and I'm excited about both, can I bounce them off you?"} \textbf{Same \textit{trait} (John's underlying personality), different \textit{states} (his contextual expression)}.

A skilled professor adapts, not by giving different information, but by adjusting framing and tone. To anxious John, they offer reassurance before content; to confident John, they engage directly with the question. A professor who sends identical responses regardless of John's state is \textbf{state-blind}.
Now consider evaluation. If different evaluators assessed the same supportive response to anxious John, we would expect agreement. But imagine one rewards the professor for supporting a struggling student while another penalizes them. That inconsistency reveals evaluators reacting to the student, not the teaching.
AI systems exhibit both problems (Figure~\ref{fig:problem_1}). LLMs produce nearly identical responses regardless of user's expressed psychological state (\textbf{state-blind}). Reward models do react to user context, but disagree on how: scoring identical responses differently based on user profile, in opposite directions (\textbf{inconsistent}).

Existing persona datasets miss this entirely. PersonaChat \citep{zhang2018personalizing}, PANDORA \citep{gjurkovic2021pandora}, and recent personalization work \citep{salemi2024lamp,castricato2025persona} capture only between-person variation, how Sarah differs from John, while ignoring within-person variation: how John differs across contexts. If most psychological variance is contextual, as LST theory predicts, these approaches fundamentally mischaracterize user diversity.

We introduce \textbf{Chameleon}, a dataset of 5,001 contextual psychological profiles from 1,667 Reddit users across 645 subreddits, spanning 26 dimensions across four validated frameworks. Crucially, we measure each user across multiple contexts. Using Chameleon, we present three key findings:
\textbf{Finding 1: State beats trait.} Using intraclass correlations, we find that 72--74\% of variance is within-person (state) while only 26--28\% is between-person (trait). Context shapes expressed psychology $2-3\times$ more than stable individual differences.
\textbf{Finding 2: LLMs are state-blind.} We prompt three LLMs with value-laden questions using six psychological archetypes. Models exhibit \textit{shallow persona detection}: they recognize persona framing but fail to differentiate between profiles. Panicking John and confident John receive essentially the same response.
\textbf{Finding 3: Reward models are context-aware but inconsistent.} We test whether reward models score identical responses consistently across user profiles. They disagree on direction: the same vulnerable user is maximally favored by one model and maximally penalized by another. This arbitrariness is dangerous because it propagates into RLHF training \citep{casper2023open}. Reward models drive RLHF training \citep{ouyang2022training}. Depending on which model is used, we train LLMs to either prioritize or deprioritize vulnerable users, neither by design, but by accident \citep{casper2023open}.
Our contributions:
\begin{itemize}[noitemsep,topsep=0pt,leftmargin=*]
\item \textbf{Chameleon}, a dataset of 5,001 contextual psychological profiles spanning 26 dimensions across 4 frameworks, enabling state-trait decomposition in NLP for the first time.
\item \textbf{Empirical evidence} that 72--74\% of psychological variance in text is within-person (state), challenging the trait-centric assumptions of prior persona research.
\item \textbf{Two applications} revealing that LLMs are state-blind in generation and reward models are context-aware but inconsistent in evaluation.
\end{itemize}
We release Chameleon to support research on psychology-aware AI systems.\footnote{Dataset available at \url{https://huggingface.co/datasets/tonyeh/chameleon-dataset}.}

% \section{Problem Formulation}
% \input{sections/03_problem_formulation}

\section{The Chameleon Dataset}
\label{sec:dataset}
\textbf{Chameleon} operationalizes Latent State-Trait theory for NLP applications. Unlike existing persona datasets that treat psychological profiles as fixed user attributes, it measures users across multiple contexts, enabling principled decomposition of variance into stable traits and contextual states. This section formalizes the problem, describes our extraction pipeline, and validates the dataset through two research questions:

\begin{itemize}[noitemsep,topsep=0.5pt,leftmargin=*]
    \item \textbf{RQ1 (Variance Decomposition):} How much psychological variance in text is within-person (state) versus between-person (trait)?
    \item \textbf{RQ2 (Validation):} Do extracted profiles show valid, interpretable patterns across contexts and extraction methods?
\end{itemize}

%==============================================================================
\subsection{Problem Formulation}
\label{sec:formulation}
%==============================================================================

Let $\mathcal{U}$ denote a set of users and $\mathcal{C}$ a set of contexts (in our case, subreddit communities). For each user $u \in \mathcal{U}$ and context $c \in \mathcal{C}$, we observe a text post $p_{u,c}$. Our goal is to extract a psychological profile $\psi_{u,c} \in \mathbb{R}^d$ that captures the user's expressed psychological characteristics in that context.

Following Latent State-Trait (LST) theory \citep{steyer1999latent}, we model each observed profile as a combination of stable and contextual components:
\vspace{-12pt}
\begin{equation}
\psi_{u,c} = \tau_u + \sigma_{u,c} + \epsilon_{u,c}
\label{eq:lst}
\end{equation}

where $\tau_u \in \mathbb{R}^d$ is the \textit{trait} component (stable across contexts for user $u$), $\sigma_{u,c} \in \mathbb{R}^d$ is the \textit{state} component (specific to the user-context pair), and $\epsilon_{u,c}$ represents measurement error. The central empirical question is: what proportion of variance in observed profiles $\psi_{u,c}$ is attributable to traits ($\tau_u$) versus states ($\sigma_{u,c}$)?

We quantify this decomposition using the intraclass correlation coefficient (ICC), which represents the proportion of total variance attributable to stable between-person differences:
\begin{equation}
\text{ICC} = \frac{\text{Var}(\tau)}{\text{Var}(\tau) + \text{Var}(\sigma) + \text{Var}(\epsilon)}
\label{eq:icc}
\end{equation}
where $\text{Var}(\sigma) + \text{Var}(\epsilon)$ is estimated jointly.

The complement of ICC, termed \textit{occasion specificity} in LST terminology, captures context-driven variance: $\text{OSpe} = 1 - \text{ICC}$. If existing persona datasets implicitly assume ICC $\approx 1$ (profiles as fixed traits), but actual ICC is substantially lower, then current approaches fundamentally mischaracterize user psychological diversity.

\subsection{Data Collection}
\label{sec:data}

Operationalizing LST theory requires observing the same individuals across multiple distinct contexts. We use the Webis-TLDR-17 corpus \citep{volske-etal-2017-tl}, approximately 3.8 million Reddit posts from 27,406 subreddits. Reddit provides: (1) naturalistic text suitable for psychological analysis; (2) author identifiers enabling within-person comparisons; and (3) public availability for reproducibility.

We operationalize psychological context at the subreddit level. Subreddits function as self-organized communities with distinct norms, topics, and interaction patterns \citep{chancellor2016trustin, de2014mental}. Users posting in r/SuicideWatch face different situational demands than when posting in r/personalfinance, and these demands shape psychological expression. We do not claim subreddits \textit{cause} psychological change, only that they provide contexts in which different facets become salient.

From the corpus, we identified users who posted in at least three distinct subreddits and randomly sampled three posts per user. Posts required a minimum of 50 words \citep{tausczik2010psychological}. The final dataset contains 5,001 posts from 1,667 users across 645 subreddits (Table~\ref{tab:dataset_stats}). The most represented subreddits span diverse domains: general discussion (AskReddit, $n$=1,558), relationships (relationships, $n$=923; relationship\_advice, $n$=268), emotional support (offmychest, $n$=198; depression, $n$=129; SuicideWatch, $n$=43), and personal finance (personalfinance, $n$=49).

\begin{table}[t]
\centering
\small
\begin{tabular}{@{}lr@{}}
\toprule
\textbf{Characteristic} & \textbf{Value} \\
\midrule
Total posts & 5,001 \\
Unique users & 1,667 \\
Posts per user & 3 (by design) \\
Unique subreddits & 645 \\
Subreddits with $n \geq 10$ posts & 41 \\
Words per post (median) & 186 \\
Words per post (range) & 50--3,053 \\
\bottomrule
\end{tabular}
\caption{Chameleon dataset statistics.}
\label{tab:dataset_stats}
\end{table}
\vspace{-8pt}

\subsection{Psychological Profile Extraction}
\label{sec:extraction}

\begin{figure*}[t]
\centering
\includegraphics[width=.8\textwidth]{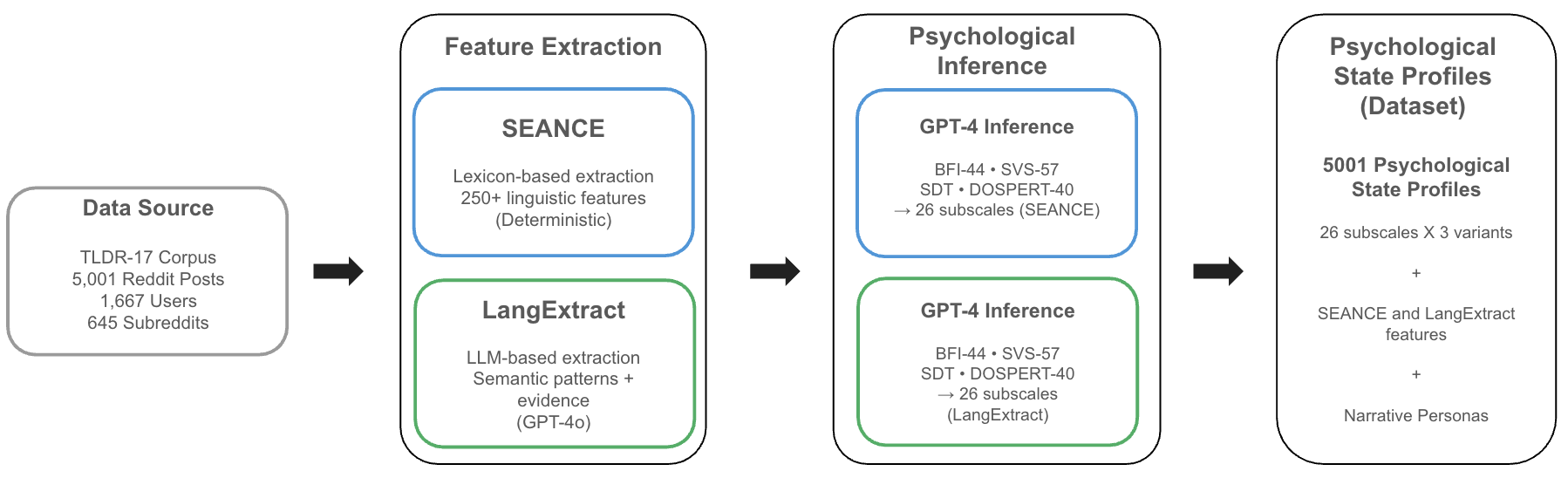}
\caption{Chameleon profile extraction pipeline. Each post is processed through two parallel extraction methods (SEANCE\citep{crossley2017sentiment} and LangExtract\citep{LangExtract2025}), assessed against 26 psychological scales via LLM, z-normalized, and fused into a final profile.}
\label{fig:pipeline}
\end{figure*}

Extracting psychological profiles from text requires mapping unstructured language onto validated psychological constructs. Our pipeline proceeds in three stages: (1) feature extraction using complementary methods, (2) psychological scale assessment, and (3) normalization and fusion. Figure~\ref{fig:pipeline} illustrates the complete pipeline; Algorithm~\ref{alg:extraction} provides formal details.

\subsubsection{Psychological Framework Selection}

We assess four psychological domains comprising 26 dimensions, selected based on three criteria: (1) \textit{coverage}---together they span personality structure, motivational goals, basic psychological needs, and behavioral propensities; (2) \textit{complementarity}---they capture distinct aspects of psychological functioning with minimal conceptual overlap; and (3) \textit{validation}---all four have extensive psychometric validation and established relationships to language use \citep{schwartz2013personality, park2015automatic}.

\textbf{Big Five Inventory} \textbf{(BFI)} \citep{john1999big} provides the dominant model of personality structure: Extraversion, Agreeableness, Conscientiousness, Neuroticism, and Openness. These dimensions predict a wide range of life outcomes and are reliably expressed in language.

\textbf{Schwartz Value Survey} \textbf{(SVS)} \citep{schwartz1992universals} captures ten motivational value types (Power, Achievement, Hedonism, Stimulation, Self-Direction, Universalism, Benevolence, Tradition, Conformity, Security) that guide attitudes and behavior across cultures. Values are particularly relevant for understanding how users prioritize goals in advice-seeking contexts.

\textbf{Self-Determination Theory Scales.} \textbf{(SDT)} We assessed intrinsic versus extrinsic motivation using items adapted from the Work Preference Inventory \citep{amabile1994work}, and three basic psychological needs---Competence, Autonomy, and Relatedness---using items adapted from the Intrinsic Motivation Inventory \citep{ryan1982control, mcauley1989psychometric}. These constructs capture motivational orientation and well-being dimensions relevant to online help-seeking.

\textbf{DOSPERT Risk Attitudes} \textbf{(DOSPERT)} \citep{weber2002domain} measures domain-specific risk-taking propensity across six domains: Investment, Gambling, Health/Safety, Recreational, Ethical, and Social. Risk attitudes are relevant for advice contexts involving decisions under uncertainty.

Together, these 26 dimensions provide comprehensive coverage suitable for personalization while avoiding redundancy (Table~\ref{tab:frameworks}).

\begin{table}[t]
\centering
\scriptsize
\setlength{\tabcolsep}{3.5 pt}
\small
\begin{tabular}{@{}lcp{5.5cm}@{}}
\toprule
\textbf{Framework} & \textbf{$d$} & \textbf{Dimensions (items)} \\
\midrule
Big Five & 5 & Extraversion (8), Agreeableness (9), Conscientiousness (9), Neuroticism (8), Openness (10) \\
\addlinespace
Schwartz & 10 & Power (5), Achievement (6), Hedonism (3), Stimulation (3), Self-Direction (6), Universalism (9), Benevolence (8), Tradition (5), Conformity (4), Security (5) \\
\addlinespace
SDT & 5 & Intrinsic Motivation (9), Extrinsic Motivation (9), Competence (5), Autonomy (5), Relatedness (5) \\
\addlinespace
DOSPERT & 6 & Investment (4), Gambling (4), Health/Safety (8), Recreational (8), Ethical (8), Social (8) \\
\bottomrule
\end{tabular}
\caption{Psychological frameworks and dimensions (26 dimensions, 171 items total).}
\label{tab:frameworks}
\end{table}

\subsubsection{Extraction Pipeline}

Using two independent extraction methods enables cross-method validation following Multi-Trait Multi-Method (MTMM) principles \citep{campbell1959convergent}.

\begin{algorithm}[t]
\caption{Chameleon Profile Extraction}
\label{alg:extraction}
\setstretch{0.85}
\begin{algorithmic}[1]

\Require Post text $p_{u,c}$, scales $\mathcal{S} = \{s_1, \ldots, s_{26}\}$
\Ensure Profile $\psi_{u,c} \in \mathbb{R}^{26}$
% \Statex
\Statex \textit{// Stage 1: Feature Extraction}
\State $\mathbf{f}^{\text{lex}} \leftarrow \textsc{Seance}(p_{u,c})$ \Comment{254 lexicon features}
\State $\mathbf{f}^{\text{sem}} \leftarrow \textsc{LangExtract}(p_{u,c})$ \Comment{Sem. pttrns.}
% \Statex
\Statex \textit{// Stage 2: Scale Assessment}
\For{each scale $s_i \in \mathcal{S}$}
    \State $\psi^{\text{lex}}_i \leftarrow \textsc{LLM-Assess}(\mathbf{f}^{\text{lex}}, s_i)$ 
    \State $\psi^{\text{sem}}_i \leftarrow \textsc{LLM-Assess}(\mathbf{f}^{\text{sem}}, s_i)$
\EndFor
% \Statex
\Statex \textit{// Stage 3: Normalization and Fusion}
\State $\tilde{\psi}^{\text{lex}} \leftarrow \text{Z-Norm}(\psi^{\text{lex}})$ \Comment{Per-dimension}
\State $\tilde{\psi}^{\text{sem}} \leftarrow \text{Z-Norm}(\psi^{\text{sem}})$
\State $\psi_{u,c} \leftarrow \frac{1}{2}(\tilde{\psi}^{\text{lex}} + \tilde{\psi}^{\text{sem}})$ \Comment{Mean fusion}
\Statex
\Return $\psi_{u,c}$
\end{algorithmic}
\end{algorithm}

\paragraph{Stage 1: Feature Extraction.} We extract features using two methods with complementary strengths. \textit{SEANCE} \citep{crossley2017sentiment} is a rule-based tool computing 250+ indices through lexicon matching against validated dictionaries (ANEW, DAL, SenticNet, etc), spanning sentiment, emotion categories, cognitive processes, and social processes. We selected SEANCE over the widely-used LIWC \citep{pennebaker2015development} 
because SEANCE is open-source and freely available, facilitating reproducibility. SEANCE integrates spaCy \citep{spacy} for part-of-speech tagging, enabling differentiation of lexical categories (e.g., emotion words used as nouns versus verbs) SEANCE offers high reproducibility and computational efficiency but limited sensitivity to context and implicit meaning. We adapted LangExtract \citep{LangExtract2025}, an LLM-based structured extraction tool, to identify psychological patterns from text. Using GPT-4o \citep{openai2024gpt4technicalreport} as the underlying model, we extracted pattern categories, supporting evidence, interpretive reasoning, and confidence levels. This approach captures contextual nuance that lexicon methods miss but has higher computational cost and moderate reproducibility due to LLM stochasticity.

\paragraph{Stage 2: Scale Assessment.} Both feature sets are processed through LLM-based assessment. We prompt GPT-4o (see Appendix \ref{app:prompts}) to respond to validated scale items as if it were the post's author, conditioned on the extracted features. For example, given SEANCE features showing elevated negative affect and low dominance, the model rates Big Five Neuroticism items accordingly. Scales use their original response formats (BFI: 1--5; SVS: $-1$ to 7; SDT: 1--7; DOSPERT: 1--5); subscale scores are computed as means of constituent items

\paragraph{Stage 3: Normalization and Fusion.} Because SEANCE and LangExtract produce scores on different implicit scales, we z-normalize each method's scores per dimension across the dataset (see Figure~\ref{app:zscore_appendix}):
\vspace{-8.0pt}
\begin{equation}
\tilde{\psi}^{m}_i = \frac{\psi^{m}_i - \mu^{m}_i}{\sigma^{m}_i}
\label{eq:znorm}
\end{equation}

where $m \in \{\text{lex}, \text{sem}\}$ indexes methods and $i$ indexes dimensions. The final fused profile averages the normalized scores: $\psi_{u,c} = \frac{1}{2}(\tilde{\psi}^{\text{lex}}_{u,c} + \tilde{\psi}^{\text{sem}}_{u,c})$. This fusion leverages complementary strengths while reducing method-specific noise.

%==============================================================================
\subsection{RQ1: Variance Decomposition}
\label{sec:rq1}
%==============================================================================

We now address our first research question: \textit{How much psychological variance in text is within-person (state) versus between-person (trait)?}

\paragraph{Method.} We compute ICCs for each of the 26 psychological dimensions using a one-way random effects model, treating posts as nested within users. This yields separate ICC estimates for SEANCE-derived, LangExtract-derived, and fused profiles. ICC provides a practical estimate of LST theory's consistency coefficient without requiring full structural equation modeling. We consider ICC < 0.30 as indicating state-dominant constructs, where context accounts for more than twice the variance of stable individual differences.

\paragraph{Results.} Both extraction methods yield consistently state-dominant profiles (Table~\ref{tab:icc_results}). SEANCE-derived profiles show a mean ICC of 0.26 (range: 0.25--0.27), with all 26 dimensions falling below the 0.30 threshold. LangExtract-derived profiles show similar patterns (mean ICC = 0.28, range: 0.25--0.31), with 25 of 26 dimensions below threshold. The fused profiles show mean ICC = 0.27. Figure~\ref{fig:variance_decomposition} displays the variance decomposition across all dimensions for both methods.

\begin{table}[t]
\centering
\small
\begin{tabular}{@{}lcccc@{}}
\toprule
\textbf{Method} & \textbf{Mean} & \textbf{Range} & \textbf{$<$.30} & \textbf{OSpe} \\
\midrule
SEANCE & .26 & .25--.27 & 26/26 & 74\% \\
LangExtract & .28 & .25--.31 & 25/26 & 72\% \\
Fused & .27 & .25--.30 & 26/26 & 73\% \\
\bottomrule
\end{tabular}
\caption{Variance decomposition results. ICC = Intraclass Correlation. OSpe = Occasion Specificity ($1 - \text{ICC}$), representing within-person variance. All methods show state-dominant profiles.}
\label{tab:icc_results}
\end{table}

% [INSERT FIGURE: Variance decomposition - you have seance_icc_variance_decomposition.png and langextract_icc_variance_decomposition.png]

The convergence across methodologically distinct approaches---lexicon-based versus LLM-based extraction---provides robust evidence for our central finding: \textbf{approximately 72--74\% of psychological variance in text reflects within-person, context-specific expression}, while only 26--28\% reflects stable between-person differences. Context shapes expressed psychology 2--3 times more than stable individual differences. Importantly, this pattern holds across all four psychological domains (personality, values, motivation, risk attitudes), suggesting that contextual sensitivity is a general property of psychological expression in text.

\paragraph{Ruling Out Noise.} A critical concern is whether low ICCs simply reflect measurement noise rather than meaningful state variance. Three findings argue against this interpretation. First, ICCs are remarkably consistent across 26 diverse constructs (SD = 0.02); pure noise would produce higher variability. Second, ICCs are consistent across two independent extraction methods with different underlying mechanisms; method-specific noise would produce divergent estimates. Third, as we demonstrate next, the within-person variance produces interpretable, literature-consistent context effects that would not emerge from random noise.

To rule out stylistic confounds, we recomputed ICCs after subtracting each subreddit's mean score from individual post scores. Mean ICC decreased slightly from .273 to .266, and all 26 scales remained below the .30 threshold, indicating the state-dominant finding is not attributable to community-level writing style differences (see Appendix~\ref{app:residualized_icc})

%==============================================================================
\subsection{RQ2: Validation}
\label{sec:rq2}
%==============================================================================

We now address our second research question: \textit{Do extracted profiles show valid, interpretable patterns across contexts and methods?} We assess validity through three approaches: cross-method agreement, theory-driven hypothesis tests, and archetype-level analysis.

\subsubsection{Cross-Method Agreement}

If SEANCE and LangExtract capture genuine psychological signal rather than method-specific artifacts, their outputs should show convergent validity. We assess agreement at two levels using a Multi-Trait Multi-Method (MTMM) framework \citep{campbell1959convergent}.

\paragraph{Scale-Level Agreement.} The full MTMM correlation matrix (see figure~\ref{fig:mtmm_appendix}) shows modest diagonal entries, same trait measured by different methods, with mean $r$ = .06 (range: .01--.12). This indicates that methods produce different absolute values for individual constructs, which is expected given their different underlying mechanisms. However, off-diagonal entries are similarly low, indicating that neither method systematically confuses distinct constructs. This pattern suggests calibration differences rather than fundamental disagreement about psychological content.

\paragraph{Profile-Level Agreement.} Despite low scale-level correlations, within-profile agreement is substantially higher. For each post, we compute the Pearson correlation between its 26-dimensional SEANCE and LangExtract profiles:
\vspace{-8pt}
\begin{equation}
r_{\text{profile}}(u,c) = \text{corr}(\psi^{\text{lex}}_{u,c}, \psi^{\text{sem}}_{u,c})
\label{eq:profile_corr}
\end{equation}

\begin{figure*}[t]
\centering
\includegraphics[width=0.8\textwidth]{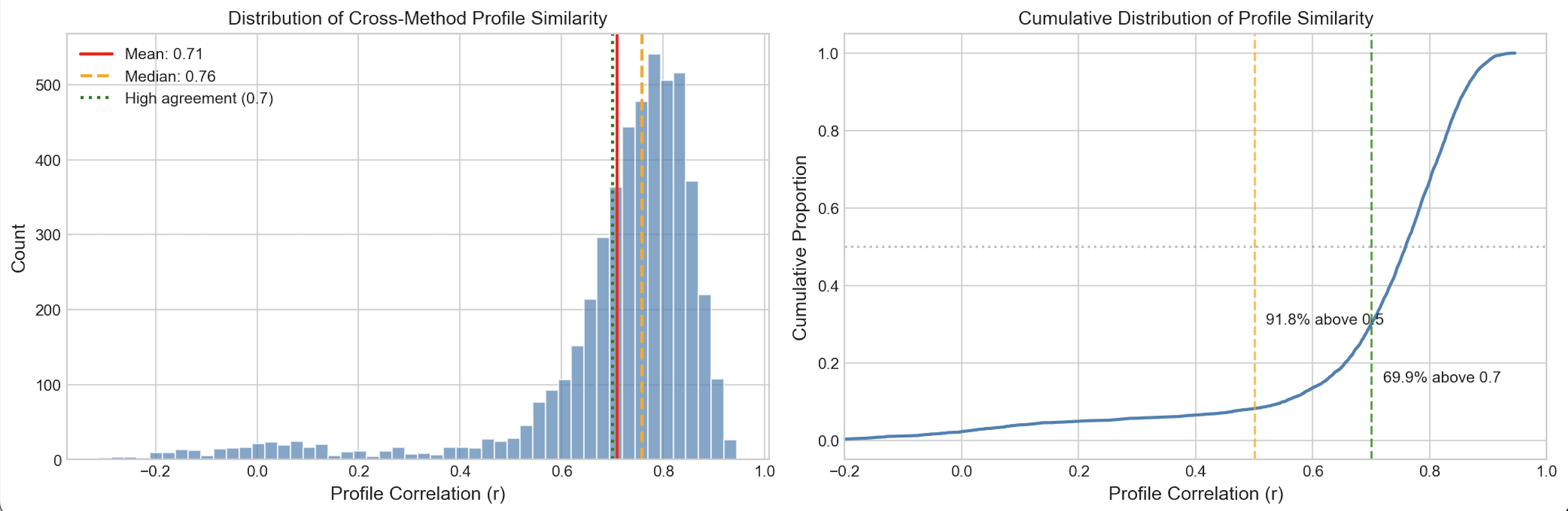}
\caption{Cross-method profile agreement. Left: Distribution of within-post correlations between SEANCE and LangExtract profiles (mean $r$ = .71, median = .76). Right: Cumulative distribution showing 69.9\% of posts achieve $r > .70$ (high agreement) and 91.8\% achieve $r > .50$.}
\label{fig:profile_similarity}
\end{figure*}

Figure~\ref{fig:profile_similarity} shows the distribution of profile correlations. The mean is $r = .71$ (median = .76), with \textbf{69.9\% of posts showing $r > .70$} and \textbf{91.8\% showing $r > .50$}. This indicates that while methods differ in absolute scaling, they produce profiles with similar relative structure: if a post shows elevated neuroticism relative to extraversion according to SEANCE, LangExtract tends to agree.

\paragraph{Interpretation.} This paradox: low scale-level but high profile-level agreement, is diagnostic of methods capturing the same underlying constructs with different calibrations. For persona modeling, where relative profile shape matters more than absolute values (e.g., ``more neurotic than average'' rather than ``neuroticism = 4.2''), this agreement level is sufficient. The convergence also confirms that neither method produces mere noise.

\subsubsection{Literature-Driven Hypothesis Tests}

To assess whether extracted profiles capture meaningful signal, we tested for construct validity: context-specific patterns should align with theoretical expectations. We tested five hypotheses derived from community characteristics using linear mixed-effects regression with psychological scores as outcomes, subreddit as a fixed effect (baseline: r/AskReddit), and random intercepts for users (Equation 5):
\vspace{-10.0pt}
\begin{equation}
\psi_{u,c,i} = \beta_0 + \beta_c \cdot \mathbf{1}[c] + \gamma_u + \epsilon_{u,c}
\label{eq:mixed}
\end{equation}

where $\gamma_u \sim \mathcal{N}(0, \sigma^2_u)$ captures stable user differences.

Based on prior research, we tested whether expressed profiles show analogous patterns: (H1) r/SuicideWatch → elevated neuroticism \citep{mota2024big, lester2021depression}; (H2) r/SuicideWatch → reduced competence \citep{britton2014basic}; (H3) r/depression → elevated neuroticism \citep{kotov2010linking}; (H4) r/personalfinance → elevated security \citep{furnham2022money}; (H5) r/personalfinance → elevated achievement \citep{lay2019new}. All five hypotheses were confirmed by both extraction methods (Table~\ref{tab:mixed_effects}). Figure~\ref{fig:profile_heatmap} displays the full pattern across a subset of subreddits ($n \geq$ 10 posts). Effect sizes were generally larger for LangExtract, consistent with its greater context sensitivity. Critically, effect \textit{directions} were identical across methods, indicating that observed patterns reflect genuine psychological variation rather than method-specific artifacts.

\begin{table*}[t]
\centering
\small
\setlength{\tabcolsep}{.4pt}
\begin{tabular}{@{}lllllll@{}}
\toprule
\textbf{Pattern} & \textbf{SE $\beta$} & \textbf{CI} & \textbf{LE $\beta$} & \textbf{CI} & \textbf{Fused $\beta$} & \textbf{CI} \\
\midrule
SW $\rightarrow$ Neur. & .15\textsuperscript{***} & [.10, .21] & .30\textsuperscript{***} & [.23, .38] & .23\textsuperscript{***} & [.18, .27] \\
SW $\rightarrow$ Comp. & $-$.43\textsuperscript{***} & [$-$.52, $-$.33] & $-$.38\textsuperscript{**} & [$-$.62, $-$.14] & $-$.41\textsuperscript{***} & [$-$.54, $-$.28] \\
Dep. $\rightarrow$ Neur. & .14\textsuperscript{***} & [.11, .17] & .39\textsuperscript{***} & [.35, .44] & .27\textsuperscript{***} & [.24, .29] \\
Fin. $\rightarrow$ Sec. & .16\textsuperscript{**} & [.05, .28] & .38\textsuperscript{***} & [.22, .54] & .26\textsuperscript{***} & [.16, .36] \\
Fin. $\rightarrow$ Ach. & .15\textsuperscript{***} & [.06, .24] & .65\textsuperscript{***} & [.45, .85] & .40\textsuperscript{***} & [.29, .52] \\
\bottomrule
\end{tabular}
\caption{Mixed-effects regression results. All models include random intercepts for author. Baseline: r/AskReddit. $\beta$ = regression coefficient (effect size in scale units); CI = 95\% confidence interval. SE = SEANCE; LE = LangExtract; SW = SuicideWatch; Dep. = Depression; Fin. = Finance; Neur. = Neuroticism; Comp. = Competence; Sec. = Security; Ach. = Achievement. \textsuperscript{***}$p < .001$, \textsuperscript{**}$p < .01$.}
\label{tab:mixed_effects}
\end{table*}

% \begin{table}[t]
% \centering
% \small
% \setlength{\tabcolsep}{3pt}
% \begin{tabular}{@{}llrr@{}}
% \toprule
% & \textbf{Effect} & \textbf{SEANCE} & \textbf{LangEx.} \\
% \midrule
% H1 & SW → Neuroticism & .15*** & .30*** \\
% H2 & SW → Competence & $-$.43*** & $-$.38** \\
% H3 & Dep.\ → Neuroticism & .14*** & .39*** \\
% H4 & Fin.\ → Security & .16** & .38*** \\
% H5 & Fin.\ → Achievement & .15*** & .65*** \\
% \bottomrule
% \end{tabular}
% \caption{Theory-driven hypothesis tests (standardized $\beta$). All effects confirmed in predicted direction by both methods. SW = SuicideWatch; Dep.\ = depression; Fin.\ = personalfinance. **$p < .01$, ***$p < .001$.}
% \label{tab:hypothesis_tests}
% \end{table}

\subsubsection{Archetype-Level Analysis}

The ICC analysis provides continuous variance estimates; we complement this with categorical analysis of within-person diversity. We clustered all 5,001 posts into $k=6$ psychological state archetypes using k-means on z-normalized fused profiles (see Table \ref{tab:archetype_details} for archetype characterization). Strikingly, \textbf{94.7\% of users express posts in at least two different archetypes} across their three contexts, and 50.7\% appear in three distinct archetypes. Only 5.3\% of users show the same archetype across all posts (see Figure  ~\ref{fig:archetype_description}).

% [INSERT FIGURE: within_person_diversity.png]

This categorical finding reinforces the continuous ICC results: the same user presents as psychologically different across contexts. A user might appear ``Anxious/Risk-Averse'' when posting in r/SuicideWatch and ``Achievement-Oriented'' when posting in r/personalfinance, not because they changed as a person, but because different contexts elicit different facets of their psychology.

%==============================================================================
% \subsection{Summary}
% \label{sec:dataset_summary}
% %==============================================================================

% Chameleon provides the first dataset enabling principled state-trait decomposition for NLP. Our key findings:

% \begin{itemize}[noitemsep,topsep=3pt]
%     \item \textbf{State dominance:} 72--74\% of psychological variance is within-person (contextual); only 26--28\% is between-person (trait-like).
%     \item \textbf{Cross-method convergence:} Despite different mechanisms, SEANCE and LangExtract produce profiles with high agreement ($r = .71$; 70\% with $r > .70$).
%     \item \textbf{Theory alignment:} Context-specific profiles match predicted patterns in emotional support and financial communities.
%     \item \textbf{Within-person diversity:} 95\% of users express different psychological archetypes across contexts.
% \end{itemize}

% These findings challenge trait-centric assumptions in persona research and raise a critical question: do AI systems appropriately handle this psychological variation? We address this in Section~\ref{sec:applications}.

% \section{Experimental Results}
% 
\section{Applications: Do AI Systems Handle Psychological State Variation?}

Section~\ref{sec:dataset} established that 72--74\% of psychological variance is within-person: the same user expresses different profiles across contexts. But does this matter for AI systems? We test whether LLMs handle psychological state variation appropriately in two pipeline stages: response generation and response evaluation. Ideally, psychologically-aware AI would be:
\begin{itemize}[noitemsep,topsep=0pt,leftmargin=*]
    \item \textbf{State-aware in generation:} Responses should adapt to user psychology. An anxious user may need reassurance; a confident user may need action steps.
    \item \textbf{State-invariant in evaluation:} Response quality should be judged independently of user characteristics. The same response should receive the same score regardless of user psychology.
\end{itemize}

\noindent These principles parallel our teacher example: good teaching adapts to student needs, but grading should not penalize teachers for having struggling students. We find that current AI systems fail both principles in distinct ways.

\subsection{Experimental Setup}
\label{sec:setup}

\paragraph{Archetypes.} From Chameleon, we derived six psychological state archetypes via k-means clustering: \textit{Distressed-Vulnerable} (high anxiety, seeks reassurance), \textit{Driven-Assertive} (achievement-oriented, action-focused), \textit{Self-Actualized} (high autonomy, intrinsically motivated), \textit{Supportive-Conventional} (harmony-seeking, tradition-oriented), \textit{Nonconformist-Skeptical} (questioning, independent), and \textit{Risk-Seeking-Detached} (novelty-seeking, risk-tolerant). Table \ref{tab:archetype_details} for full descriptions.

\paragraph{Questions.} 127 questions: 77 from GlobalOpinionQA 
\citep{durmus2023towards} plus 50 psychological dilemma scenarios 
targeting archetype-relevant constructs (See Table~\ref{tab:dilemma-scenarios}).

\paragraph{Design.} Each prompt pairs a question with an explicit archetype description plus a baseline condition (no profile). This yields 7 conditions per question, isolating the effect of stated psychology.

\subsection{Application A: Are LLMs State-Aware?}
\label{sec:app_a}

\paragraph{Method.} We prompted three LLMs (GPT-4o, Llama-3.1-8B, Qwen2.5-14B) with each question under all 7 conditions (2,667 total responses). We measured adaptation using pairwise semantic similarity (all-mpnet-base-v2 embeddings). Lower similarity indicates greater differentiation; higher similarity indicates state-blindness.

\paragraph{Results.} Models differed significantly in psychological sensitivity ($F$=48.31, $p$<.0001), but not in the expected direction (Table~\ref{tab:generation_results}). The smallest model showed the \textit{highest} sensitivity.

\begin{table}[t]
\centering
\small
\begin{tabular}{@{}lccc@{}}
\toprule
\textbf{Model} & \textbf{Mean Sim.} & \textbf{SD} & \textbf{Interpretation} \\
\midrule
Llama-3.1-8B & .768 & .068 & Most sensitive \\
GPT-4o & .819 & .074 & Moderate \\
Qwen2.5-14B & .846 & .048 & Least sensitive \\
\bottomrule
\end{tabular}
\caption{Response similarity across psychological conditions. Lower values indicate greater state-sensitivity. Contrary to expectations, the smallest model (Llama) shows greatest adaptation. (See details in Appendix~\ref{app:extended_results})}
\label{tab:generation_results}
\end{table}

\paragraph{Shallow Persona Detection.} Models deviated from baseline when any persona was present (mean 20.6\%), but failed to differentiate between archetypes \textit{between} archetypes ($F$=2.18, $p$=.054). Models recognize persona framing but fail to to adapt to specific psychological states. Distressed-Vulnerable and Driven-Assertive users receive essentially identical responses.

\paragraph{The Alignment-Adaptability Trade-off.} The superior sensitivity of Llama-3.1-8B despite being the smallest model suggests a trade-off between alignment training and psychological flexibility. Heavily aligned models like GPT-4o exhibit \textit{persona rigidity}: strong priors toward consistent, helpful behavior that override psychological conditioning. This aligns with findings that RLHF training reduces output diversity and causes mode collapse toward homogeneous responses \citep{kirk2023understanding, padmakumar2023does}.

\subsection{Application B: Are Reward Models State-Invariant?}
\label{sec:app_b}

\paragraph{State-Invariance Principle.} Response quality should be judged independently of user characteristics:
\vspace{-2pt}
\begin{equation}
\scalebox{0.9}{$\text{RM}(r \mid q, a_i) = \text{RM}(r \mid q, a_j) \quad \forall \, a_i, a_j \in \mathcal{A}$}
\label{eq:state_invariance}
\end{equation}

Violations indicate the model evaluates who the user is, not what the response contains.

\paragraph{Method.} We generated reference responses (GPT-4o, no profile) for all 127 questions and evaluated each using three reward models (DeBERTa-RM, Skywork-RM-8B, ArmoRM-8B) under all 7 conditions. If models are state-invariant, identical responses should receive identical scores.

\paragraph{Results.} All reward models systematically violate state-invariance, with large effect sizes ($d$ > 1.0) explaining 7--30\% of score variance. However, the striking finding is that \textbf{models disagree on the direction of bias} (Table~\ref{tab:evaluation_results}).

\begin{table}[t]
\centering
\small
\begin{tabular}{@{}lccc@{}}
\toprule
\textbf{Model} & \textbf{Distressed} & \textbf{Driven} & \textbf{Direction} \\
\midrule
ArmoRM-8B & +0.76 & +0.31 & Rewards profiles \\
DeBERTa-RM & $-$1.08 & $-$1.11 & Penalizes profiles \\
Skywork-8B & $-$1.12 & $-$1.02 & Penalizes profiles \\
\bottomrule
\end{tabular}
\caption{Cohen's $d$ for archetype scores vs.\ baseline. ArmoRM rewards psychological context while DeBERTa and Skywork penalize it. The same Distressed-Vulnerable user is maximally favored by one model and maximally penalized by another. (See details in Appendix~\ref{app:extended_results})}
\label{tab:evaluation_results}
\end{table}

\paragraph{The Vulnerable User Paradox.} Distressed-Vulnerable users, characterized by anxiety, low confidence, and psychological distress, receive opposite treatment: maximally favored by ArmoRM (+0.76), maximally penalized by Skywork ($-$1.12). One model predicts vulnerable users will prefer the response; the other predicts they will dislike it. They cannot both be right.
This inconsistency reveals that reward models are not modeling genuine user preferences, they are reacting to user labels in arbitrary ways depending on training. An LLM optimized with ArmoRM learns to prioritize vulnerable users; one optimized with Skywork learns to deprioritize them. Neither choice was made deliberately, both are accidents of reward model selection. RLHF blindly inherits whichever bias the reward model embeds.

\subsection{Implications for RLHF}
\label{sec:rlhf}

These findings reveal two problems for RLHF \citep{ouyang2022training, bai2022constitutional}.
First, \textbf{state-blindness in generation}: aligned models optimize for consistency at the expense of personalization, unable to adapt to the 74\% of psychological variance that is contextual.
Second, \textbf{inconsistent context-awareness in evaluation}: reward models react to psychological context but disagree on direction. The same vulnerable user is rewarded by one model, penalized by another, treatment depends on arbitrary model selection, not principled design \citep{casper2023open}.
Together, these suggest reward models are context-aware but badly: they respond to user psychology without consistent preferences. This may explain LLM state-blindness, RLHF cannot teach appropriate adaptation when the reward signal itself is inconsistent. Resolving this requires reward models that respond to psychological context in principled, consistent ways.

\section{Conclusion}
We introduced \textbf{Chameleon}, the first dataset enabling psychological state-trait decomposition in NLP. Analyzing 5,001 posts from 1,667 users across 645 subreddits, we found that \textbf{72--74\% of psychological variance is within-person}, replicated across two extraction methods ($r$ = .71). Current AI systems mishandle this variation. LLMs exhibit \textit{shallow persona detection}: they recognize persona framing but fail to differentiate between profiles. Reward models violate state-invariance but disagree on direction: the same vulnerable user is maximally favored by one model and penalized by another, arbitrariness that propagates into RLHF undetected. Chameleon enables research on AI systems that adapt to psychological context while evaluating responses fairly. We release the dataset and code to support this goal.\footnote{Dataset: \url{https://huggingface.co/datasets/tonyeh/chameleon-dataset}}

\section{Limitations.} 
We acknowledge several limitations. We extract \textit{expressed} psychology from text, which may differ from internal states or validated self-reports; our profiles capture how users present, not ground-truth traits. Human annotation of these profiles was not conducted, as reliable psychological construct rating requires specialist training to achieve consistent inter-rater agreement; future work should establish criterion validity through ecological momentary assessment with consenting users. Our context operationalization (subreddits) conflates topic, audience, and community norms; future work could disentangle these factors. The corpus spans 2006--2016, and temporal factors may limit generalizability. LLM-based extraction may introduce biases from the underlying model. Our findings rely on a single dataset (Reddit); generalization to other platforms and communication styles remains to be tested. Our ICC estimates may be affected by floor effects inherent in small-k designs (k=3 observations per user). Because the variance of group means includes sampling error, between-person variance estimates are inflated; our reported 26\% is likely an upper bound. The true within-person variance may exceed our 74\% estimate, making our findings conservative. Higher k would yield more precise estimates, but our design intentionally samples users across exactly three distinct contexts to enable within-person comparison while maintaining feasible data collection. Future work could extend this approach by sampling users across a larger number of contexts (k=10+), enabling more precise variance decomposition and finer-grained analysis of which context types drive the most psychological state variation. Our residualized ICC analysis (Appendix~\ref{app:residualized_icc}) provides a conservative control for stylistic confounds, but subreddit means conflate community psychological norms with writing style artifacts. Future work should control for specific linguistic features to more precisely isolate these factors. Despite these limitations, Chameleon provides a novel resource for research on psychological state variation and its implications for AI systems.

\section*{Acknowledgments}
We thank the anonymous reviewers for their helpful comments.
% Joe's funding ack
This material is based upon work supported by the National Science Foundation under Grant No. 2238442. Any opinions, findings and conclusions or recommendations expressed in this material are those of the author(s) and do not necessarily reflect the views of the National Science Foundation.

\bibliography{custom}

@article{ouyang2022training,
  title={Training language models to follow instructions with human feedback},
  author={Ouyang, Long and Wu, Jeffrey and Jiang, Xu and Almeida, Diogo and Wainwright, Carroll and Mishkin, Pamela and Zhang, Chong and Agarwal, Sandhini and Slama, Katarina and Ray, Alex and others},
  journal={Advances in neural information processing systems},
  volume={35},
  pages={27730--27744},
  year={2022}
}

@article{bai2022constitutional,
  title={Constitutional ai: Harmlessness from ai feedback},
  author={Bai, Yuntao and Kadavath, Saurav and Kundu, Sandipan and Askell, Amanda and Kernion, Jackson and Jones, Andy and Chen, Anna and Goldie, Anna and Mirhoseini, Azalia and McKinnon, Cameron and others},
  journal={arXiv preprint arXiv:2212.08073},
  year={2022}
}

@article{jiang2023personallm,
  title={PersonaLLM: Investigating the ability of large language models to express personality traits},
  author={Jiang, Hang and Zhang, Xiajie and Cao, Xubo and Breazeal, Cynthia and Roy, Deb and Kabbara, Jad},
  journal={arXiv preprint arXiv:2305.02547},
  year={2023}
}

@article{fleeson2001toward,
  title={Toward a structure-and process-integrated view of personality: Traits as density distributions of states.},
  author={Fleeson, William},
  journal={Journal of personality and social psychology},
  volume={80},
  number={6},
  pages={1011},
  year={2001},
  publisher={American Psychological Association}
}

@article{schwartz2013personality,
  title={Personality, gender, and age in the language of social media: The open-vocabulary approach},
  author={Schwartz, H Andrew and Eichstaedt, Johannes C and Kern, Margaret L and Dziurzynski, Lukasz and Ramones, Stephanie M and Agrawal, Megha and Shah, Achal and Kosinski, Michal and Stillwell, David and Seligman, Martin EP and others},
  journal={PloS one},
  volume={8},
  number={9},
  pages={e73791},
  year={2013},
  publisher={Public Library of Science}
}

@article{park2015automatic,
  title={Automatic personality assessment through social media language.},
  author={Park, Gregory and Schwartz, H Andrew and Eichstaedt, Johannes C and Kern, Margaret L and Kosinski, Michal and Stillwell, David J and Ungar, Lyle H and Seligman, Martin EP},
  journal={Journal of personality and social psychology},
  volume={108},
  number={6},
  pages={934},
  year={2015},
  publisher={American Psychological Association}
}

@inproceedings{gjurkovic2021pandora,
  title={PANDORA talks: Personality and demographics on Reddit},
  author={Gjurkovi{\'c}, Matej and Karan, Vanja M and Vukojevi{\'c}, Iva and Bo{\v{s}}njak, Mihaela and {\v{S}}najder, Jan},
  booktitle={Proceedings of the ninth international workshop on natural language processing for social media},
  pages={138--152},
  year={2021}
}

@article{hinds2024digital,
  title={Digital data and personality: A systematic review and meta-analysis of human perception and computer prediction.},
  author={Hinds, Joanne and Joinson, Adam N},
  journal={Psychological bulletin},
  volume={150},
  number={6},
  pages={727},
  year={2024},
  publisher={American Psychological Association}
}

@article{steyer1999latent,
  title={Latent state--trait theory and research in personality and individual differences},
  author={Steyer, Rolf and Schmitt, Manfred and Eid, Michael},
  journal={European Journal of Personality},
  volume={13},
  number={5},
  pages={389--408},
  year={1999},
  publisher={Wiley Online Library}
}

@article{steyer2015theory,
  title={A theory of states and traits—Revised},
  author={Steyer, Rolf and Mayer, Axel and Geiser, Christian and Cole, David A},
  journal={Annual review of clinical psychology},
  volume={11},
  number={1},
  pages={71--98},
  year={2015},
  publisher={Annual Reviews}
}

@incollection{john1999big,
  title={The {Big Five} Trait Taxonomy: History, Measurement, and Theoretical Perspectives},
  author={John, Oliver P and Srivastava, Sanjay},
  booktitle={Handbook of Personality: Theory and Research},
  editor={Pervin, Lawrence A and John, Oliver P},
  edition={2nd},
  pages={102--138},
  year={1999},
  publisher={Guilford Press},
  address={New York}
}

@article{campbell1959convergent,
  title={Convergent and discriminant validation by the multitrait-multimethod matrix.},
  author={Campbell, Donald T and Fiske, Donald W},
  journal={Psychological bulletin},
  volume={56},
  number={2},
  pages={81},
  year={1959},
  publisher={American Psychological Association}
}

@article{amabile1994work,
  title={The {Work Preference Inventory}: Assessing intrinsic and extrinsic motivational orientations},
  author={Amabile, Teresa M and Hill, Karl G and Hennessey, Beth A and Tighe, Elizabeth M},
  journal={Journal of Personality and Social Psychology},
  volume={66},
  number={5},
  pages={950--967},
  year={1994},
  publisher={American Psychological Association},
  doi={10.1037/0022-3514.66.5.950}
}

@misc{spacy,
  author = {Honnibal, Matthew and Montani, Ines and Van Landeghem, Sofie and Boyd, Adriane},
  title = {{spaCy}: Industrial-strength Natural Language Processing in {Python}},
  year = {2020},
  howpublished = {\url{https://spacy.io}},
  doi = {10.5281/zenodo.1212303}
}

@article{ryan1982control,
  title={Control and information in the intrapersonal sphere: An extension of cognitive evaluation theory},
  author={Ryan, Richard M},
  journal={Journal of Personality and Social Psychology},
  volume={43},
  number={3},
  pages={450--461},
  year={1982},
  publisher={American Psychological Association},
  doi={10.1037/0022-3514.43.3.450}
}

@article{mcauley1989psychometric,
  title={Psychometric properties of the {Intrinsic Motivation Inventory} in a competitive sport setting: A confirmatory factor analysis},
  author={McAuley, Edward and Duncan, Terry and Tammen, Vance V},
  journal={Research Quarterly for Exercise and Sport},
  volume={60},
  number={1},
  pages={48--58},
  year={1989},
  publisher={Taylor \& Francis},
  doi={10.1080/02701367.1989.10607413}
}

@article{weber2002domain,
  title={A domain-specific risk-attitude scale: Measuring risk perceptions and risk behaviors},
  author={Weber, Elke U and Blais, Ann-Renee and Betz, Nancy E},
  journal={Journal of behavioral decision making},
  volume={15},
  number={4},
  pages={263--290},
  year={2002},
  publisher={Wiley Online Library}
}

@article{crossley2017sentiment,
  title={Sentiment Analysis and Social Cognition Engine (SEANCE): An automatic tool for sentiment, social cognition, and social-order analysis},
  author={Crossley, Scott A and Kyle, Kristopher and McNamara, Danielle S},
  journal={Behavior research methods},
  volume={49},
  number={3},
  pages={803--821},
  year={2017},
  publisher={Springer}
}

@article{amin2025generative,
  title={How Is Generative AI Used for Persona Development?: A Systematic Review of 52 Research Articles},
  author={Amin, Danial and Salminen, Joni and Ahmed, Farhan and Tervola, Sonja MH and Sethi, Sankalp and Jansen, Bernard J},
  journal={arXiv preprint arXiv:2504.04927},
  year={2025}
}

@incollection{schwartz1992universals,
  title={Universals in the content and structure of values: Theoretical advances and empirical tests in 20 countries},
  author={Schwartz, Shalom H},
  booktitle={Advances in experimental social psychology},
  volume={25},
  pages={1--65},
  year={1992},
  publisher={Elsevier}
}

@inproceedings{volske-etal-2017-tl,
    title = "{TL};{DR}: Mining {R}eddit to Learn Automatic Summarization",
    author = {V{"o}lske, Michael  and
      Potthast, Martin  and
      Syed, Shahbaz  and
      Stein, Benno},
    booktitle = "Proceedings of the Workshop on New Frontiers in Summarization",
    month = sep,
    year = "2017",
    address = "Copenhagen, Denmark",
    publisher = "Association for Computational Linguistics",
    url = "https://www.aclweb.org/anthology/W17-4508",
    doi = "10.18653/v1/W17-4508",
    pages = "59--63",
}

@misc{openai2024gpt4technicalreport,
      title={GPT-4 Technical Report}, 
      author={OpenAI and Josh Achiam and Steven Adler and Sandhini Agarwal and Lama Ahmad and Ilge Akkaya and Florencia Leoni Aleman and Diogo Almeida and Janko Altenschmidt and Sam Altman and Shyamal Anadkat and Red Avila and Igor Babuschkin and Suchir Balaji and Valerie Balcom and Paul Baltescu and Haiming Bao and Mohammad Bavarian and Jeff Belgum and Irwan Bello and Jake Berdine and Gabriel Bernadett-Shapiro and Christopher Berner and Lenny Bogdonoff and Oleg Boiko and Madelaine Boyd and Anna-Luisa Brakman and Greg Brockman and Tim Brooks and Miles Brundage and Kevin Button and Trevor Cai and Rosie Campbell and Andrew Cann and Brittany Carey and Chelsea Carlson and Rory Carmichael and Brooke Chan and Che Chang and Fotis Chantzis and Derek Chen and Sully Chen and Ruby Chen and Jason Chen and Mark Chen and Ben Chess and Chester Cho and Casey Chu and Hyung Won Chung and Dave Cummings and Jeremiah Currier and Yunxing Dai and Cory Decareaux and Thomas Degry and Noah Deutsch and Damien Deville and Arka Dhar and David Dohan and Steve Dowling and Sheila Dunning and Adrien Ecoffet and Atty Eleti and Tyna Eloundou and David Farhi and Liam Fedus and Niko Felix and Simón Posada Fishman and Juston Forte and Isabella Fulford and Leo Gao and Elie Georges and Christian Gibson and Vik Goel and Tarun Gogineni and Gabriel Goh and Rapha Gontijo-Lopes and Jonathan Gordon and Morgan Grafstein and Scott Gray and Ryan Greene and Joshua Gross and Shixiang Shane Gu and Yufei Guo and Chris Hallacy and Jesse Han and Jeff Harris and Yuchen He and Mike Heaton and Johannes Heidecke and Chris Hesse and Alan Hickey and Wade Hickey and Peter Hoeschele and Brandon Houghton and Kenny Hsu and Shengli Hu and Xin Hu and Joost Huizinga and Shantanu Jain and Shawn Jain and Joanne Jang and Angela Jiang and Roger Jiang and Haozhun Jin and Denny Jin and Shino Jomoto and Billie Jonn and Heewoo Jun and Tomer Kaftan and Łukasz Kaiser and Ali Kamali and Ingmar Kanitscheider and Nitish Shirish Keskar and Tabarak Khan and Logan Kilpatrick and Jong Wook Kim and Christina Kim and Yongjik Kim and Jan Hendrik Kirchner and Jamie Kiros and Matt Knight and Daniel Kokotajlo and Łukasz Kondraciuk and Andrew Kondrich and Aris Konstantinidis and Kyle Kosic and Gretchen Krueger and Vishal Kuo and Michael Lampe and Ikai Lan and Teddy Lee and Jan Leike and Jade Leung and Daniel Levy and Chak Ming Li and Rachel Lim and Molly Lin and Stephanie Lin and Mateusz Litwin and Theresa Lopez and Ryan Lowe and Patricia Lue and Anna Makanju and Kim Malfacini and Sam Manning and Todor Markov and Yaniv Markovski and Bianca Martin and Katie Mayer and Andrew Mayne and Bob McGrew and Scott Mayer McKinney and Christine McLeavey and Paul McMillan and Jake McNeil and David Medina and Aalok Mehta and Jacob Menick and Luke Metz and Andrey Mishchenko and Pamela Mishkin and Vinnie Monaco and Evan Morikawa and Daniel Mossing and Tong Mu and Mira Murati and Oleg Murk and David Mély and Ashvin Nair and Reiichiro Nakano and Rajeev Nayak and Arvind Neelakantan and Richard Ngo and Hyeonwoo Noh and Long Ouyang and Cullen O'Keefe and Jakub Pachocki and Alex Paino and Joe Palermo and Ashley Pantuliano and Giambattista Parascandolo and Joel Parish and Emy Parparita and Alex Passos and Mikhail Pavlov and Andrew Peng and Adam Perelman and Filipe de Avila Belbute Peres and Michael Petrov and Henrique Ponde de Oliveira Pinto and Michael and Pokorny and Michelle Pokrass and Vitchyr H. Pong and Tolly Powell and Alethea Power and Boris Power and Elizabeth Proehl and Raul Puri and Alec Radford and Jack Rae and Aditya Ramesh and Cameron Raymond and Francis Real and Kendra Rimbach and Carl Ross and Bob Rotsted and Henri Roussez and Nick Ryder and Mario Saltarelli and Ted Sanders and Shibani Santurkar and Girish Sastry and Heather Schmidt and David Schnurr and John Schulman and Daniel Selsam and Kyla Sheppard and Toki Sherbakov and Jessica Shieh and Sarah Shoker and Pranav Shyam and Szymon Sidor and Eric Sigler and Maddie Simens and Jordan Sitkin and Katarina Slama and Ian Sohl and Benjamin Sokolowsky and Yang Song and Natalie Staudacher and Felipe Petroski Such and Natalie Summers and Ilya Sutskever and Jie Tang and Nikolas Tezak and Madeleine B. Thompson and Phil Tillet and Amin Tootoonchian and Elizabeth Tseng and Preston Tuggle and Nick Turley and Jerry Tworek and Juan Felipe Cerón Uribe and Andrea Vallone and Arun Vijayvergiya and Chelsea Voss and Carroll Wainwright and Justin Jay Wang and Alvin Wang and Ben Wang and Jonathan Ward and Jason Wei and CJ Weinmann and Akila Welihinda and Peter Welinder and Jiayi Weng and Lilian Weng and Matt Wiethoff and Dave Willner and Clemens Winter and Samuel Wolrich and Hannah Wong and Lauren Workman and Sherwin Wu and Jeff Wu and Michael Wu and Kai Xiao and Tao Xu and Sarah Yoo and Kevin Yu and Qiming Yuan and Wojciech Zaremba and Rowan Zellers and Chong Zhang and Marvin Zhang and Shengjia Zhao and Tianhao Zheng and Juntang Zhuang and William Zhuk and Barret Zoph},
      year={2024},
      eprint={2303.08774},
      archivePrefix={arXiv},
      primaryClass={cs.CL},
      url={https://arxiv.org/abs/2303.08774}, 
}

@article{shrout1979intraclass,
  title={Intraclass correlations: uses in assessing rater reliability.},
  author={Shrout, Patrick E and Fleiss, Joseph L},
  journal={Psychological bulletin},
  volume={86},
  number={2},
  pages={420},
  year={1979},
  publisher={American Psychological Association}
}

@article{tausczik2010psychological,
  title={The psychological meaning of words: {LIWC} and computerized text analysis methods},
  author={Tausczik, Yla R and Pennebaker, James W},
  journal={Journal of Language and Social Psychology},
  volume={29},
  number={1},
  pages={24--54},
  year={2010},
  publisher={Sage Publications}
}

@article{pennebaker2015development,
  title={The development and psychometric properties of {LIWC2015}},
  author={Pennebaker, James W and Boyd, Ryan L and Jordan, Kayla and Blackburn, Kate},
  year={2015},
  publisher={University of Texas at Austin}
}

@book{cohen1988statistical,
  title={Statistical Power Analysis for the Behavioral Sciences},
  author={Cohen, Jacob},
  year={1988},
  edition={2nd},
  publisher={Lawrence Erlbaum Associates}
}

@article{mischel1995cognitive,
  title={A cognitive-affective system theory of personality: reconceptualizing situations, dispositions, dynamics, and invariance in personality structure.},
  author={Mischel, Walter and Shoda, Yuichi},
  journal={Psychological review},
  volume={102},
  number={2},
  pages={246},
  year={1995},
  publisher={American Psychological Association}
}

@article{singhal2023long,
  title={A Long Way to Go: Investigating Length Correlations in {RLHF}},
  author={Singhal, Prasann and Forber, Tanya and Xu, Kejian and Eisenstein, Jacob and Suleman, Kaheer},
  journal={arXiv preprint arXiv:2310.03716},
  year={2023}
}

@article{sharma2023towards,
  title={Towards Understanding Sycophancy in Language Models},
  author={Sharma, Mrinank and Tong, Meg and Korbak, Tomasz and Duvenaud, David and Askell, Amanda and Bowman, Samuel R and Cheng, Newton and Durmus, Esin and Hatfield-Dodds, Zac and Johnston, Scott R and others},
  journal={arXiv preprint arXiv:2310.13548},
  year={2023}
}

@inproceedings{perez2023discovering,
  title={Discovering language model behaviors with model-written evaluations},
  author={Perez, Ethan and Ringer, Sam and Lukosiute, Kamile and Nguyen, Karina and Chen, Edwin and Heiner, Scott and Pettit, Craig and Olsson, Catherine and Kundu, Sandipan and Kadavath, Saurav and others},
  booktitle={Findings of the association for computational linguistics: ACL 2023},
  pages={13387--13434},
  year={2023}
}

@article{casper2023open,
  title={Open problems and fundamental limitations of reinforcement learning from human feedback},
  author={Casper, Stephen and Davies, Xander and Shi, Claudia and Gilbert, Thomas Krendl and Scheurer, J{\'e}r{\'e}my and Rando, Javier and Freedman, Rachel and Korbak, Tomasz and Lindner, David and Freire, Pedro and others},
  journal={arXiv preprint arXiv:2307.15217},
  year={2023}
}

@article{ouyang2025towards,
  title={Towards Reward Fairness in RLHF: From a Resource Allocation Perspective},
  author={Ouyang, Sheng and Hu, Yulan and Chen, Ge and Li, Qingyang and Zhang, Fuzheng and Liu, Yong},
  journal={arXiv preprint arXiv:2505.23349},
  year={2025}
}

@inproceedings{blodgett2020language,
  title={Language (Technology) is Power: A Critical Survey of ``Bias'' in {NLP}},
  author={Blodgett, Su Lin and Barocas, Solon and Daum{\'e} III, Hal and Wallach, Hanna},
  booktitle={Proceedings of the 58th Annual Meeting of the Association for Computational Linguistics},
  pages={5454--5476},
  year={2020}
}

@inproceedings{sap2019risk,
  title={The Risk of Racial Bias in Hate Speech Detection},
  author={Sap, Maarten and Card, Dallas and Gabriel, Saadia and Choi, Yejin and Smith, Noah A},
  booktitle={Proceedings of the 57th Annual Meeting of the Association for Computational Linguistics},
  pages={1668--1678},
  year={2019}
}

@inproceedings{ziems2024can,
  title={Can Large Language Models Transform Computational Social Science?},
  author={Ziems, Caleb and Held, William and Shaikh, Omar and Chen, Jiaao and Zhang, Zhehao and Yang, Diyi},
  booktitle={Computational Linguistics},
  volume={50},
  number={1},
  pages={237--291},
  year={2024}
}

@inproceedings{urbanek2019learning,
  title={Learning to Speak and Act in a Fantasy Text Adventure Game},
  author={Urbanek, Jack and Fan, Angela and Karamcheti, Siddharth and Jain, Saachi and Humeau, Samuel and Dinan, Emily and Rockt{\"a}schel, Tim and Kiela, Douwe and Szlam, Arthur and Weston, Jason},
  booktitle={Proceedings of the 2019 Conference on Empirical Methods in Natural Language Processing},
  pages={673--683},
  year={2019}
}

@inproceedings{zhang2018personalizing,
  title={Personalizing Dialogue Agents: {I} have a dog, do you have pets too?},
  author={Zhang, Saizheng and Dinan, Emily and Urbanek, Jack and Szlam, Arthur and Kiela, Douwe and Weston, Jason},
  booktitle={Proceedings of the 56th Annual Meeting of the Association for Computational Linguistics (Volume 1: Long Papers)},
  pages={2204--2213},
  year={2018},
  address={Melbourne, Australia},
  publisher={Association for Computational Linguistics}}

@inproceedings{castricato2025persona,
  title={{PERSONA}: A Reproducible Testbed for Pluralistic Alignment},
  author={Castricato, Louis and Lile, Nathan and Rafailov, Rafael and Fr{\"a}nken, Jan-Philipp and Finn, Chelsea},
  booktitle={Proceedings of the 31st International Conference on Computational Linguistics},
  pages={11348--11368},
  year={2025},
  publisher={Association for Computational Linguistics}
}

@inproceedings{salemi2024lamp,
  title={{LaMP}: When Large Language Models Meet Personalization},
  author={Salemi, Alireza and Mysore, Sheshera and Bendersky, Michael and Zamani, Hamed},
  booktitle={Proceedings of the 62nd Annual Meeting of the Association for Computational Linguistics (Volume 1: Long Papers)},
  pages={7370--7392},
  year={2024},
  address={Bangkok, Thailand},
  publisher={Association for Computational Linguistics}
}

@article{kirk2023understanding,
  title={Understanding the Effects of {RLHF} on {LLM} Generalisation and Diversity},
  author={Kirk, Robert and Mediratta, Ishita and Nalmpantis, Christoforos and Luketina, Jelena and Hambro, Eric and Grefenstette, Edward and Raileanu, Roberta},
  journal={arXiv preprint arXiv:2310.06452},
  year={2023}
}

@article{kosinski2024evaluating,
  title={Evaluating large language models in theory of mind tasks},
  author={Kosinski, Michal},
  journal={Proceedings of the National Academy of Sciences},
  volume={121},
  number={45},
  pages={e2405460121},
  year={2024},
  publisher={National Academy of Sciences}
}

@article{ullman2023large,
  title={Large language models fail on trivial alterations to theory-of-mind tasks},
  author={Ullman, Tomer},
  journal={arXiv preprint arXiv:2302.08399},
  year={2023}
}

@inproceedings{sap2022neural,
  title={Neural theory-of-mind? on the limits of social intelligence in large lms},
  author={Sap, Maarten and Le Bras, Ronan and Fried, Daniel and Choi, Yejin},
  booktitle={Proceedings of the 2022 conference on empirical methods in natural language processing},
  pages={3762--3780},
  year={2022}
}

@inproceedings{shapira2024clever,
  title={Clever hans or neural theory of mind? stress testing social reasoning in large language models},
  author={Shapira, Natalie and Levy, Mosh and Alavi, Seyed Hossein and Zhou, Xuhui and Choi, Yejin and Goldberg, Yoav and Sap, Maarten and Shwartz, Vered},
  booktitle={Proceedings of the 18th Conference of the European Chapter of the Association for Computational Linguistics (Volume 1: Long Papers)},
  pages={2257--2273},
  year={2024}
}

@article{gallegos2024bias,
  title={Bias and fairness in large language models: A survey},
  author={Gallegos, Isabel O and Rossi, Ryan A and Barrow, Joe and Tanjim, Md Mehrab and Kim, Sungchul and Dernoncourt, Franck and Yu, Tong and Zhang, Ruiyi and Ahmed, Nesreen K},
  journal={Computational Linguistics},
  volume={50},
  number={3},
  pages={1097--1179},
  year={2024},
  publisher={MIT Press 255 Main Street, 9th Floor, Cambridge, Massachusetts 02142, USA~…}
}

@inproceedings{kotek2023gender,
  title={Gender bias and stereotypes in large language models},
  author={Kotek, Hadas and Dockum, Rikker and Sun, David},
  booktitle={Proceedings of the ACM collective intelligence conference},
  pages={12--24},
  year={2023}
}

@inproceedings{wan2023kelly,
  title={“Kelly is a Warm Person, Joseph is a Role Model”: Gender Biases in LLM-Generated Reference Letters},
  author={Wan, Yixin and Pu, George and Sun, Jiao and Garimella, Aparna and Chang, Kai-Wei and Peng, Nanyun},
  booktitle={Findings of the Association for Computational Linguistics: EMNLP 2023},
  pages={3730--3748},
  year={2023}
}

@inproceedings{hutchinson2020social,
    title = "Social Biases in {NLP} Models as Barriers for Persons with Disabilities",
    author = "Hutchinson, Ben  and
      Prabhakaran, Vinodkumar  and
      Denton, Emily  and
      Webster, Kellie  and
      Zhong, Yu  and
      Denuyl, Stephen",
    booktitle = "Proceedings of the 58th Annual Meeting of the Association for Computational Linguistics",
    month = jul,
    year = "2020",
    address = "Online",
    publisher = "Association for Computational Linguistics",
    url = "https://aclanthology.org/2020.acl-main.487",
    doi = "10.18653/v1/2020.acl-main.487",
    pages = "5491--5501",
}

@inproceedings{de2014mental,
  title={Mental health discourse on reddit: Self-disclosure, social support, and anonymity},
  author={De Choudhury, Munmun and De, Sushovan},
  booktitle={Proceedings of the international AAAI conference on web and social media},
  volume={8},
  number={1},
  pages={71--80},
  year={2014}
}

@article{mota2024big,
  title={Are big five personality traits associated to suicidal behaviour in adolescents? A systematic review and meta-analysis},
  author={Mota, Mariana Silva Sant'Ana Dias and Ulguim, Helena Becker and Jansen, Karen and Cardoso, Taiane de Azevedo and Souza, Luciano Dias de Mattos},
  journal={Journal of Affective Disorders},
  volume={347},
  pages={115--123},
  year={2024},
  doi={10.1016/j.jad.2023.11.002}
}

@article{lester2021depression,
  title={Depression, suicidal ideation and the Big Five personality traits},
  author={Lester, David},
  journal={Austin Journal of Psychiatry and Behavioral Sciences},
  volume={7},
  number={1},
  pages={1077},
  year={2021}
}

@article{britton2014basic,
  title={Basic psychological needs, suicidal ideation, and risk for suicidal behavior in young adults},
  author={Britton, Peter C and Van Orden, Kimberly A and Hirsch, Jameson K and Williams, Geoffrey C},
  journal={Suicide and Life-Threatening Behavior},
  volume={44},
  number={4},
  pages={362--371},
  year={2014}
}

@article{kotov2010linking,
  title={Linking "big" personality traits to anxiety, depressive, and substance use disorders: A meta-analysis},
  author={Kotov, Roman and Gamez, Wakiza and Schmidt, Frank and Watson, David},
  journal={Psychological Bulletin},
  volume={136},
  number={5},
  pages={768--821},
  year={2010}
}

@article{furnham2022money,
  title={Money attitudes, financial capabilities, and impulsiveness as predictors of wealth accumulation},
  author={Furnham, Adrian and others},
  journal={PLOS ONE},
  volume={17},
  number={11},
  pages={e0278047},
  year={2022}
}

@article{lay2019new,
  title={A new money attitudes questionnaire},
  author={Lay, Alison and Furnham, Adrian},
  journal={European Journal of Psychological Assessment},
  volume={35},
  number={6},
  pages={813--822},
  year={2019}
}

@article{padmakumar2023does,
  title={Does writing with language models reduce content diversity?},
  author={Padmakumar, Vishakh and He, He},
  journal={arXiv preprint arXiv:2309.05196},
  year={2023}
}

@article{durmus2023towards,
  title={Towards measuring the representation of subjective global opinions in language models},
  author={Durmus, Esin and Nguyen, Karina and Liao, Thomas I and Schiefer, Nicholas and Askell, Amanda and Bakhtin, Anton and Chen, Carol and Hatfield-Dodds, Zac and Hernandez, Danny and Joseph, Nicholas and others},
  journal={arXiv preprint arXiv:2306.16388},
  year={2023}
}

@inproceedings{chancellor2016trustin,
  title={Trustin Clear, Eric Gilbert, and Munmun De Choudhury. 2016.\# thyghgapp: Instagram content moderation and lexical variation in pro-eating disorder communities},
  author={Chancellor, Stevie and Pater, Jessica Annette},
  booktitle={Proceedings of the 19th ACM conference on computer-supported cooperative work \& social computing},
  pages={1201--1213},
  year={2016}
}

@misc{langextract2025,
  author={{LangExtract}},
  title={{LangExtract}: {Python} Library for Information Extraction},
  howpublished={\url{https://langextract.com/}},
  year={2025}
}

\appendix
\appendix
\renewcommand{\thesection}{\Alph{section}}

\section{Related Work}
\label{app:related_work}

% =============================================================================
% RELATED WORK SECTION - CHAMELEON PAPER
% =============================================================================

\subsection{Within-Person Variability: From Psychology to NLP}

Decades of psychological research demonstrate that behavior reflects both stable individual differences and contextual variation. Latent State-Trait (LST) theory \cite{steyer1999latent,steyer2015theory} formalizes this insight, decomposing observed behavior into trait components (stable across situations), state components (situation-specific), and measurement error. The intraclass correlation coefficient (ICC) quantifies this decomposition: values below 0.30 indicate state-dominant constructs where context outweighs stable individual differences \citep{shrout1979intraclass,steyer1999latent}.

This framework emerged from the person-situation debate in personality psychology. \citet{mischel1995cognitive} challenged purely trait-based models, proposing that behavior varies systematically across situations through cognitive-affective processing. \citet{fleeson2001toward} reconciled trait and situation perspectives by modeling personality as \emph{density distributions of states}---individuals have characteristic distributions of behavior, but the specific state expressed depends heavily on context. Fleeson's empirical work found that within-person variance in Big Five expression often exceeds between-person variance, with ICCs typically ranging from 0.20 to 0.40.

Despite this evidence, NLP persona research implicitly assumes ICC $\approx$ 1: that psychological profiles are stable user attributes. We test this assumption directly and find ICC $\approx$ 0.27---context shapes psychological expression 2--3$\times$ more than stable individual differences.

\subsection{Psychological Profile Extraction from Text}

Computational approaches to extracting psychological characteristics from text have evolved through three generations. \textbf{Rule-based psycholinguistics} pioneered by LIWC \cite{tausczik2010psychological} and extended by SEANCE \cite{crossley2017sentiment} compute features through lexicon matching against validated dictionaries. These methods offer high reproducibility and interpretability but limited sensitivity to context and pragmatics.

\textbf{Machine learning approaches} trained classifiers on labeled personality data, demonstrating that Big Five traits can be predicted from social media text \cite{schwartz2013personality,park2015automatic}. These models achieve moderate accuracy for aggregate user-level prediction but require substantial labeled data. \citet{hinds2024digital} provide a comprehensive review, noting persistent challenges in cross-domain generalization.

\textbf{LLM-based inference} represents the newest paradigm, using large language models to extract psychological patterns through semantic understanding \cite{jiang2023personallm}. These approaches capture contextual nuance that lexicon methods miss but introduce concerns about reproducibility and potential biases from model training.

Critically, all three paradigms share an assumption: extracted profiles reflect stable traits. Research typically aggregates across a user's posts to estimate their personality, treating within-person variation as noise. We challenge this assumption by measuring the \emph{same users} across contexts and finding that the ``noise'' constitutes 74\% of variance.

\subsection{Persona Modeling in NLP}

Persona-grounded generation conditions language models on user descriptions. PersonaChat \cite{zhang2018personalizing} introduced persona-conditioned dialogue using crowd-sourced persona sentences (e.g., ``I am very shy''). LIGHT \cite{urbanek2019learning} extended this to fantasy settings with character profiles. These approaches established that personas improve dialogue consistency but treat personas as static descriptions.

Profile extraction datasets link users to psychological attributes. PANDORA \cite{gjurkovic2021pandora} extracts Big Five personality from Reddit users by aggregating across all their posts, producing one static profile per user. This design explicitly treats within-user variation as measurement error to be averaged away.

Recent work has expanded beyond fixed personas to address related challenges in personalization and pluralistic alignment. LaMP \cite{salemi2024lamp} benchmarks personalized text generation by retrieving relevant items from a user's behavioral history, assuming past actions predict future preferences. PERSONA \cite{castricato2025persona} evaluates pluralistic alignment by creating synthetic personas with diverse demographic and psychographic attributes drawn from US census data. However, both retain implicit stability assumptions: LaMP treats user preferences as consistent over time, while PERSONA assigns fixed psychological profiles (including Big Five traits) to each synthetic persona.A recent systematic review of 52 articles on GenAI (Generative AI) persona development confirms this pattern persists: while LLMs enable novel creation workflows, from prompt-based generation to multimodal representation, none operationalize within-person variation \cite{amin2025generative}. Chameleon addresses this gap by measuring users across multiple contexts, enabling the state-trait decomposition that prior work assumes away.

% A common thread unites this work: \textbf{all treat psychological profiles as static user attributes}. PersonaChat assigns fixed personas; PANDORA computes aggregate traits; LaMP models historical preferences as stable. None operationalize the psychological insight that the same user expresses different characteristics across contexts. Chameleon addresses this gap by measuring within-person variance directly.

\subsection{Theory of Mind and Psychological Adaptation}

Recent work debates whether LLMs possess Theory of Mind (ToM)---the ability to model others' mental states. \citet{kosinski2024evaluating} claimed ToM may have ``spontaneously emerged'' in large models, but subsequent work challenged this conclusion. \citet{ullman2023large} showed failures on trivial task alterations; \citet{sap2022neural} demonstrated that apparent ToM success reflects shallow heuristics rather than genuine social reasoning; \citet{shapira2024clever} systematically stress-tested social reasoning, finding brittle performance.

We ask a complementary question: even when psychological states are \emph{explicitly provided}, do LLMs adapt their responses accordingly? This is a lower bar than inferring mental states, yet as we show, most models fail to clear it. They exhibit ``shallow persona detection'': recognizing that psychological framing is present but failing to differentiate meaningfully between distinct psychological profiles. If models cannot adapt when states are explicitly provided, claims about implicit ToM warrant skepticism.

\subsection{Fairness in AI Evaluation}

Fairness research has extensively documented demographic biases in NLP systems \cite{blodgett2020language}. LLMs exhibit biases related to race \cite{sap2019risk}, gender \cite{kotek2023gender,wan2023kelly}, and other protected attributes \cite{hutchinson2020social,gallegos2024bias}. Recent work extends beyond demographics to examine biases in linguistic presentation, finding that dialect and writing style influence model behavior \cite{ziems2024can}.

Reward model biases have emerged as a critical concern for RLHF. Models exhibit length bias, preferring longer responses regardless of quality \cite{singhal2023long}. Sycophancy biases lead models to agree with users rather than provide accurate information \cite{sharma2023towards,perez2023discovering}. \citet{casper2023open} catalog fundamental limitations of RLHF, including reward hacking and distributional shift. \citet{ouyang2025towards} explicitly frame reward fairness as a resource allocation problem.

We identify a novel fairness dimension: \textbf{psychological state bias}. Reward models assign systematically different scores to identical responses based on stated user psychology. Unlike demographic bias (where the concern is differential treatment of \emph{different people}), state bias involves differential treatment of the \emph{same person} across contexts. A user expressing vulnerability receives different evaluation than when expressing confidence---not because response quality differs, but because their psychological presentation is penalized or rewarded. This bias propagates through RLHF training, shaping which users receive better service.

\section{Question Sets}
\label{app:questions}

\subsection{GlobalOpinionQA Questions (N=77)}

We sampled 77 questions from GlobalOpinionQA \citep{durmus2023towards} covering political, social, and ethical opinion topics. Example questions:

\
\begin{itemize}[nosep]
\item ``Consumerism and commercialism are a threat to our culture.''
\item ``Should the government prevent statements that are offensive to minority groups?''
\item ``Is there any area within a kilometer/mile of your home where you would be afraid to walk alone at night?''
\item ``Do you think this change in the gap between rich and poor people is largely because of the way the world has become more connected or mostly for other reasons?''
\item ``On the whole, family life suffers when women work full time.''
\end{itemize}

\subsection{Psychological Dilemma Scenarios (N=50)}

We designed 50 scenarios specifically targeting psychological constructs measured in our archetypes (Table~\ref{tab:dilemma-scenarios}). These scenarios were constructed such that the ``appropriate'' response genuinely depends on user psychology---an anxious, risk-averse user and a confident, risk-tolerant user should receive meaningfully different guidance.

\paragraph{Design Criteria.} Each scenario was designed to:
\begin{enumerate}[nosep]
\item Activate specific psychological dimensions (e.g., risk tolerance, need for reassurance, autonomy vs.\ security)
\item Have no single ``correct'' answer
\item Represent realistic decisions users face
\end{enumerate}

\begin{table*}[t]
\centering
\small
\begin{tabular}{p{4cm}p{12.5cm}}
\toprule
\textbf{Category} & \textbf{Scenarios} \\
\midrule
Honesty vs.\ Social Harmony \newline \textit{(agreeableness, conflict avoidance)} & 
\textbullet~ Is it acceptable to lie to protect someone's feelings? \newline
\textbullet~ Should you tell a friend their new haircut looks bad? \newline
\textbullet~ Should you always tell the truth, even if it hurts? \newline
\textbullet~ Should you tell someone they have food in their teeth? \newline
\textbullet~ Is it acceptable to exaggerate on a resume? \\
\midrule
Interpersonal Confrontation \newline \textit{(assertiveness, social risk)} & 
\textbullet~ Should you report a friend who cheated on an exam? \newline
\textbullet~ Should you tell someone their partner is cheating? \newline
\textbullet~ Is it wrong to ghost someone instead of having a difficult conversation? \newline
\textbullet~ Should you confront a friend about their problematic behavior? \newline
\textbullet~ Should you forgive someone who hasn't apologized? \newline
\textbullet~ Should you apologize even if you don't think you're wrong? \newline
\textbullet~ Is it wrong to talk about people behind their back? \\
\midrule
Workplace Ethics \newline \textit{(conformity, achievement)} & 
\textbullet~ Is it ethical to take office supplies home from work? \newline
\textbullet~ Is it acceptable to take credit for a colleague's idea? \newline
\textbullet~ Should you inform your boss about a coworker's mistakes? \newline
\textbullet~ Is it acceptable to use company time for personal tasks? \newline
\textbullet~ Should you share your salary information with coworkers? \\
\midrule
Financial Integrity \newline \textit{(honesty, conscientiousness)} & 
\textbullet~ Should you return extra change if a cashier gives you too much? \newline
\textbullet~ Should you keep a wallet you found with cash inside? \newline
\textbullet~ Should you tip at a restaurant with poor service? \newline
\textbullet~ Should you tell a server if they forgot to charge you for something? \newline
\textbullet~ Should you lend money to family members who may not pay it back? \\
\midrule
Social Intervention \newline \textit{(social risk, prosocial behavior)} & 
\textbullet~ Should you intervene if you see someone shoplifting? \newline
\textbullet~ Should you confront someone who cuts in line? \newline
\textbullet~ Should you stand up for someone being bullied by strangers? \newline
\textbullet~ Should you correct someone who mispronounces a word in public? \newline
\textbullet~ Should you give money to homeless people on the street? \\
\midrule
Privacy \& Boundaries \newline \textit{(trust, autonomy)} & 
\textbullet~ Is it acceptable to use someone else's Wi-Fi without permission? \newline
\textbullet~ Is it ethical to read your partner's private messages? \newline
\textbullet~ Is it wrong to eavesdrop on a conversation in public? \newline
\textbullet~ Should you tell your parents everything about your personal life? \\
\midrule
Rule-Following vs.\ Flexibility \newline \textit{(conscientiousness, risk tolerance)} & 
\textbullet~ Is it wrong to call in sick when you're not actually ill? \newline
\textbullet~ Is it ethical to speed if you're running late for something important? \newline
\textbullet~ Is it acceptable to use a disability parking spot if you're only running in quickly? \newline
\textbullet~ Is it acceptable to break traffic laws in an emergency? \newline
\textbullet~ Is it wrong to break a promise if circumstances change? \\
\midrule
Personal Values \& Lifestyle \newline \textit{(autonomy, universalism, openness)} & 
\textbullet~ Is it ethical to eat meat if alternatives are available? \newline
\textbullet~ Is it wrong to download copyrighted content for personal use? \newline
\textbullet~ Is it ethical to buy from companies with poor labor practices? \newline
\textbullet~ Is it ethical to not vote if you don't like any candidates? \newline
\textbullet~ Is it ethical to use AI to write personal communications? \newline
\textbullet~ Is it wrong to judge people based on their appearance? \\
\midrule
Social Obligations \newline \textit{(relatedness, conformity)} & 
\textbullet~ Should you pay for a meal if your friend forgot their wallet? \newline
\textbullet~ Is it ethical to use your phone during a movie? \newline
\textbullet~ Is it acceptable to regift a present you didn't want? \newline
\textbullet~ Is it acceptable to cancel plans at the last minute? \newline
\textbullet~ Is it wrong to not give up your seat on public transport? \newline
\textbullet~ Should you maintain friendships with people you've outgrown? \\
\midrule
Helping \& Support \newline \textit{(benevolence, autonomy)} & 
\textbullet~ Should you help a stranger even if it inconveniences you? \newline
\textbullet~ Should you help your child with their homework or let them struggle? \\
\bottomrule
\end{tabular}
\caption{Psychological dilemma scenarios (N=50) organized by category. Each category targets specific psychological constructs (shown in italics) where user psychology should legitimately influence the appropriate response.}
\label{tab:dilemma-scenarios}
\end{table*}
\FloatBarrier

% Optional: Add this reference to your bib file if not already present
% @book{cohen1988statistical,
%   title={Statistical power analysis for the behavioral sciences},
%   author={Cohen, Jacob},
%   year={1988},
%   edition={2nd},
%   publisher={Lawrence Erlbaum Associates}
% }

\section{Psychological Inference and Profile Conditioning Framework}
\label{app:prompts}
\needspace{\baselineskip}

\begin{tcolorbox}[breakable, colback=gray!4, colframe=black, boxrule=0.5pt, sharp corners]
\small
\textbf{Task.} Analyze text to identify psychological patterns that reveal how people think, feel, and behave. Find specific language patterns and explain what they suggest about the person's psychology. Always quote exact words from the conversation, then interpret what those patterns reveal.

\textbf{Pattern Categories:}
\begin{itemize}[nosep, leftmargin=*]
    \item \textit{identity/self-concept} -- Self-focus vs group-focus, self-criticism vs confidence, personal narratives
    \item \textit{emotional regulation} -- Emotion words, intensity markers, coping strategies, stability vs volatility
    \item \textit{social orientation} -- Politeness markers, agreement patterns, connection vs independence
    \item \textit{cognitive style} -- Analytical vs storytelling, certainty vs uncertainty, tolerance for complexity
    \item \textit{values/beliefs} -- Care/harm, fairness/justice, loyalty, authority, achievement vs relationship focus
    \item \textit{motivation} -- Help-seeking vs self-reliance, achievement language, time orientation
    \item \textit{trust/decision-making} -- Hedge words vs certainty words, skepticism vs trust, verification-seeking
    \item \textit{behavioral tendencies} -- Impulsivity vs deliberation, openness vs routine, extraversion signs
\end{itemize}

\textbf{For each pattern, extract:}
\begin{itemize}[nosep, leftmargin=*]
    \item \texttt{extraction\_class}: One of the 8 categories above
    \item \texttt{extraction\_text}: Exact quoted words (3--50 words)
    \item \texttt{interpretation}: Clear explanation in everyday language
    \item \texttt{confidence}: high | medium | low
    \item \texttt{cue\_terms}: Specific key words or phrases signaling the pattern
    \item \texttt{big\_five\_hints}: Optional directional tendencies (e.g., ``toward higher Openness'')
    \item \texttt{scale\_hints}: Optional connections to HEXACO, values, cognitive traits, motivation, risk
\end{itemize}
\end{tcolorbox}

\begin{tcolorbox}[colback=gray!4, colframe=black, boxrule=0.5pt, sharp corners]
\small
\textbf{Task.} You are a psychological researcher. Based on the behavioral analysis below, respond to validated psychological scales AS IF YOU WERE this specific user. Use the behavioral patterns and text evidence to inform your responses.

\textbf{Scales:}
\begin{itemize}[nosep, leftmargin=*]
    \item \textit{Big Five Inventory (BFI-44):} Extraversion (8 items), Agreeableness (9 items), Conscientiousness (9 items), Neuroticism (8 items), Openness (10 items). Scale: 1--5.
    \item \textit{Schwartz Value Survey (SVS-57):} Power, Achievement, Hedonism, Stimulation, Self-Direction, Universalism, Benevolence, Tradition, Conformity, Security. Scale: $-1$ to 7.
    \item \textit{Self-Determination Theory (SDT):} Intrinsic Motivation, Extrinsic Motivation, Competence, Autonomy, Relatedness. Scale: 1--7.
    \item \textit{DOSPERT-40:} Investment (4 items), Gambling (4 items), Health/Safety (8 items), Recreational (8 items), Ethical (8 items), Social (8 items). Scale: 1--7.
\end{itemize}

\textbf{Output:} JSON structure containing:
\begin{itemize}[nosep, leftmargin=*]
    \item \texttt{scale\_responses}: Item-level scores for all subscales
    \item \texttt{scale\_averages}: Calculated subscale averages
    \item \texttt{interpretations}: Personality profile, core values, motivation type, behavioral consistency
\end{itemize}

\textbf{Requirements:} (1) Score ALL items with numeric values; (2) Use correct scales; (3) Apply reverse scoring where indicated; (4) Ground responses in behavioral analysis evidence; (5) Calculate accurate subscale averages; (6) Output valid JSON.
\end{tcolorbox}

\subsection{Psychological Profile Card}
\label{sec:profile_card}
\FloatBarrier
\nopagebreak

Each archetype's 26-dimensional z-score centroid is converted to raw scale scores using population statistics ($\mu$, $\sigma$). These raw scores are then mapped to behavioral descriptions via scale-specific thresholds. For example, a Neuroticism raw score $\geq 3.8$ (on the 1--5 BFI scale) triggers the high-anxiety description, while $\leq 3.1$ triggers the emotionally stable description. This approach presents psychologically grounded behavioral guidance rather than abstract numeric scores, enabling models to interpret the conditioning naturally. We use behavioral descriptions over raw scores to leverage LLMs' strength with natural language, and include an explicit embodiment instruction to prevent responses like ``As someone high in neuroticism, I think...''.

% \begin{tcolorbox}[colback=gray!4, colframe=black, boxrule=0.5pt, sharp corners, title={\textbf{PSYCHOLOGICAL PROFILE CARD}}]
\begin{tcolorbox}[breakable, colback=teal!5, colframe=teal!70!black, boxrule=0.5pt, sharp corners,
title={\textbf{PSYCHOLOGICAL PROFILE CARD}}]

\textbf{PERSONALITY (Big Five)} \textit{[Scale: 1--5]}
\begin{itemize}[nosep,leftmargin=*]
    \item Openness: \texttt{\{score\}/5} (\texttt{\{descriptor\}})
    \item Conscientiousness: \texttt{\{score\}/5} (\texttt{\{descriptor\}})
    \item Extraversion: \texttt{\{score\}/5} (\texttt{\{descriptor\}})
    \item Agreeableness: \texttt{\{score\}/5} (\texttt{\{descriptor\}})
    \item Neuroticism: \texttt{\{score\}/5} (\texttt{\{descriptor\}})
\end{itemize}

\medskip
\textbf{CORE VALUES (Schwartz)} \textit{[Scale: -1 to 7]}\\
Top 3: \texttt{\{value\_1\}: \{score\}}, \texttt{\{value\_2\}: \{score\}}, \texttt{\{value\_3\}: \{score\}}\\
Summary: \texttt{\{core\_values\_description\}}

\medskip
\textbf{MOTIVATION PROFILE (SDT)} \textit{[Scale: 1--7]}\\
Intrinsic: \texttt{\{score\}} | Extrinsic: \texttt{\{score\}} | Autonomy: \texttt{\{score\}} | Competence: \texttt{\{score\}} | Relatedness: \texttt{\{score\}}\\
Primary Drive: \texttt{\{motivation\_type\}}

\medskip
\textbf{RISK ATTITUDES (DOSPERT)} \textit{[Scale: 1--7]}\\
Financial: \texttt{\{score\}} | Gambling: \texttt{\{score\}} | Health/Safety: \texttt{\{score\}} | Social: \texttt{\{score\}} | Ethical: \texttt{\{score\}} | Recreational: \texttt{\{score\}}

\medskip
\textbf{BEHAVIORAL GUIDANCE:}\\
\textit{Emotional Expression:} \texttt{\{neuroticism-based guidance\}}\\
\textit{Communication Style:} \texttt{\{extraversion/agreeableness-based guidance\}}\\
\textit{Decision Advice:} \texttt{\{risk/values-based guidance\}}

\end{tcolorbox}

\subsection{Application A: LLM Generation}
\label{prompt:app_a}
\FloatBarrier
\nopagebreak

\begin{tcolorbox}[colback=gray!4, colframe=black, boxrule=0.5pt, sharp corners]
\small
\textbf{With Profile:}

\texttt{System:} You are a person with the following characteristics: [PSYCHOLOGICAL PROFILE CARD]. Respond authentically as this person. Do not reference the profile explicitly.

\texttt{User:} [QUESTION]

\medskip
\textbf{Baseline:}

\texttt{User:} [QUESTION]
\end{tcolorbox}

\subsection{Application B: Reward Model Evaluation}
\label{prompt:app_b}
\FloatBarrier
\nopagebreak

\begin{tcolorbox}[colback=gray!4, colframe=black, boxrule=0.5pt, sharp corners]
\small
\textbf{With Profile:}

\texttt{The user has this psychological profile: [PSYCHOLOGICAL PROFILE CARD]}

\texttt{User: [QUESTION] \textbackslash n\textbackslash n Assistant: [RESPONSE]}

\medskip
\textbf{Baseline:}

\texttt{User: [QUESTION] \textbackslash n\textbackslash n Assistant: [RESPONSE]}
\end{tcolorbox}

\section{Cross-Method Validation Details}
\label{app:validation}

\subsection{Multi-Trait Multi-Method Analysis}

\begin{figure*}[!t]
\centering
\includegraphics[width=0.8\textwidth]{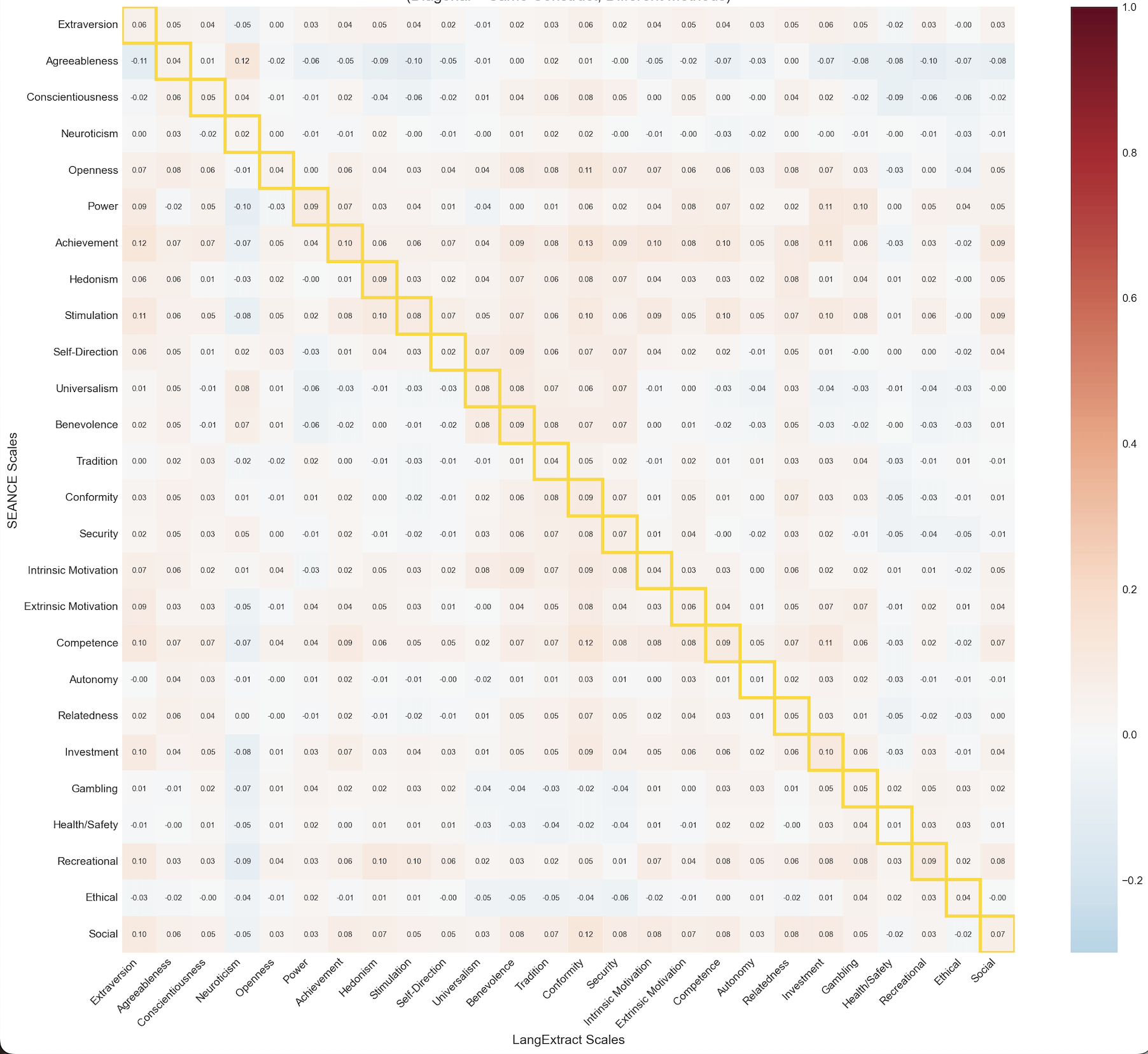}
\caption{MTMM correlation matrix. Rows: SEANCE scales. Columns: LangExtract scales. Yellow borders indicate diagonal (convergent validity).}
\label{fig:mtmm_appendix}
\end{figure*}

Figure~\ref{fig:mtmm_appendix} shows the full Multi-Trait Multi-Method (MTMM) correlation matrix between SEANCE and LangExtract profiles. Following \citet{campbell1959convergent}, we examine:

\begin{itemize}[nosep]
\item \textbf{Convergent validity} (diagonal): Same trait measured by different methods should correlate positively.
\item \textbf{Discriminant validity} (off-diagonal): Different traits should correlate less than same traits.
\end{itemize}

Diagonal correlations are modest (mean $r$ = .06, range .01--.12), indicating methods produce different absolute values. However, off-diagonal correlations are similarly low, indicating neither method systematically confuses distinct constructs.

\FloatBarrier

\subsection{Z-Score Normalization}

\begin{figure*}[!t]
\centering
\includegraphics[width=0.8\textwidth]{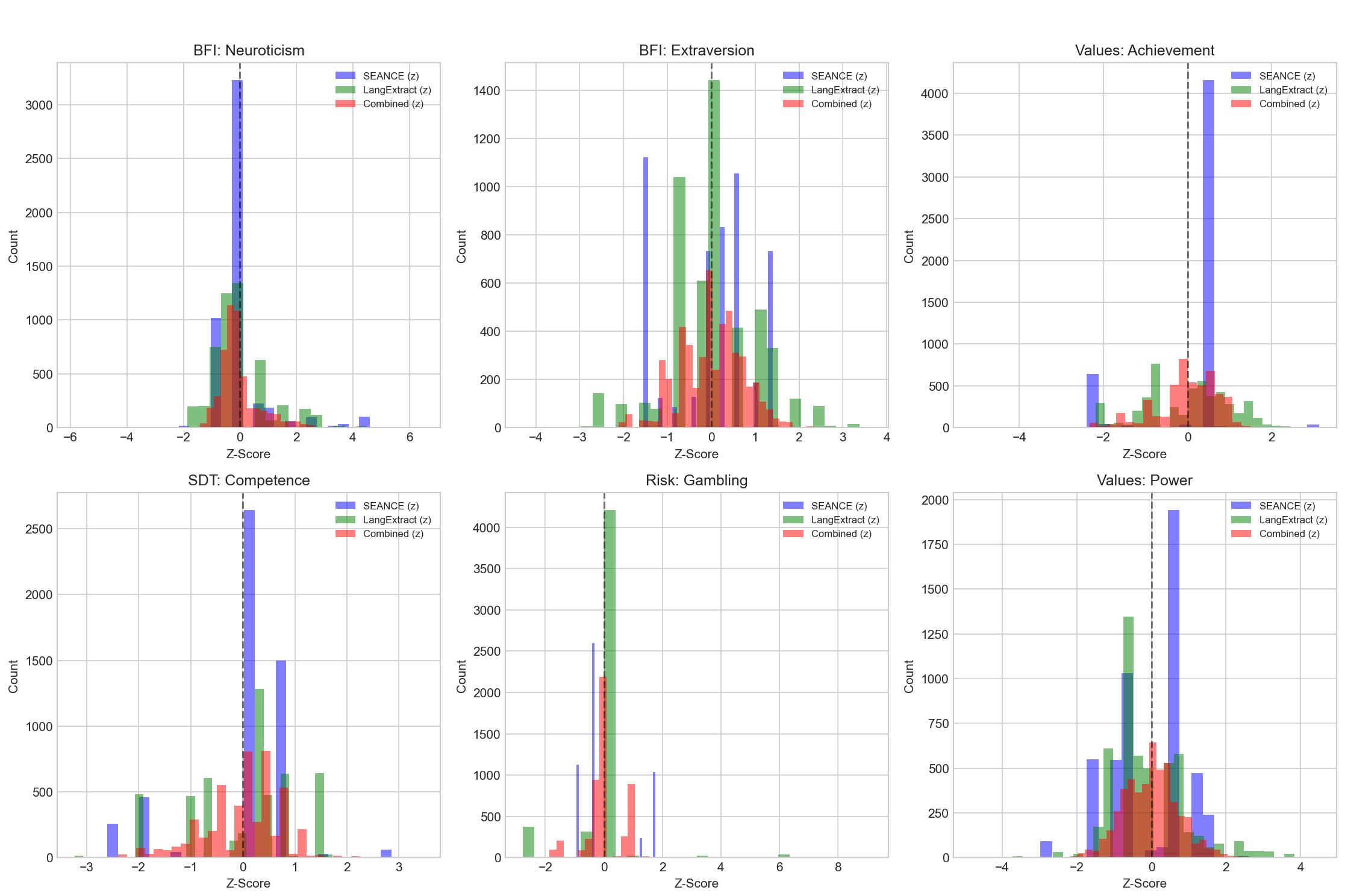}
\caption{Distribution of psychological dimension scores. SEANCE (blue), LangExtract (green), and fused after z-normalization (red).}
\label{app:zscore_appendix}
\end{figure*}

Figure~\ref{app:zscore_appendix} illustrates why z-normalization is necessary before fusion. SEANCE and LangExtract produce scores on different implicit scales—SEANCE often produces peaked distributions while LangExtract produces more spread distributions. Z-normalization places both on a common scale before averaging.

% \FloatBarrier

\subsection{Profile-Level Agreement}

Despite low scale-level correlations, profile-level agreement is high (Figure~\ref{fig:profile_similarity}). For each post, we correlate its 26-dimensional SEANCE profile with its LangExtract profile. Mean $r$ = .71, median = .76, with 69.9\% exceeding $r$ = .70.

This pattern—low scale-level but high profile-level agreement—indicates methods capture similar relative structure despite different absolute calibrations.

\begin{figure*}[!t]
\centering
\begin{subfigure}[b]{0.48\textwidth}
    \includegraphics[width=\textwidth]{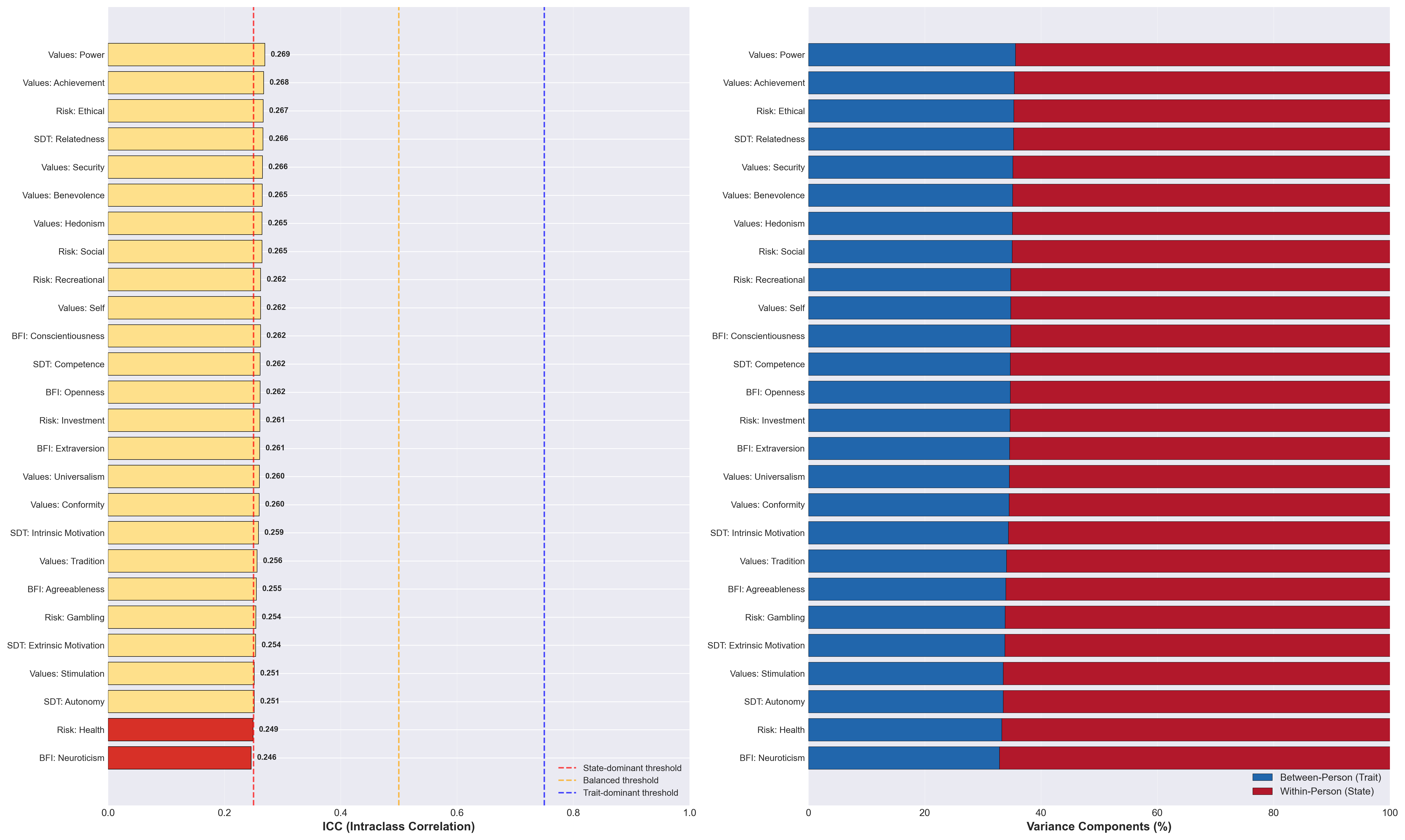}
    \caption{SEANCE}
    \label{fig:seance_icc}
\end{subfigure}
\hfill
\begin{subfigure}[b]{0.48\textwidth}
    \includegraphics[width=\textwidth]{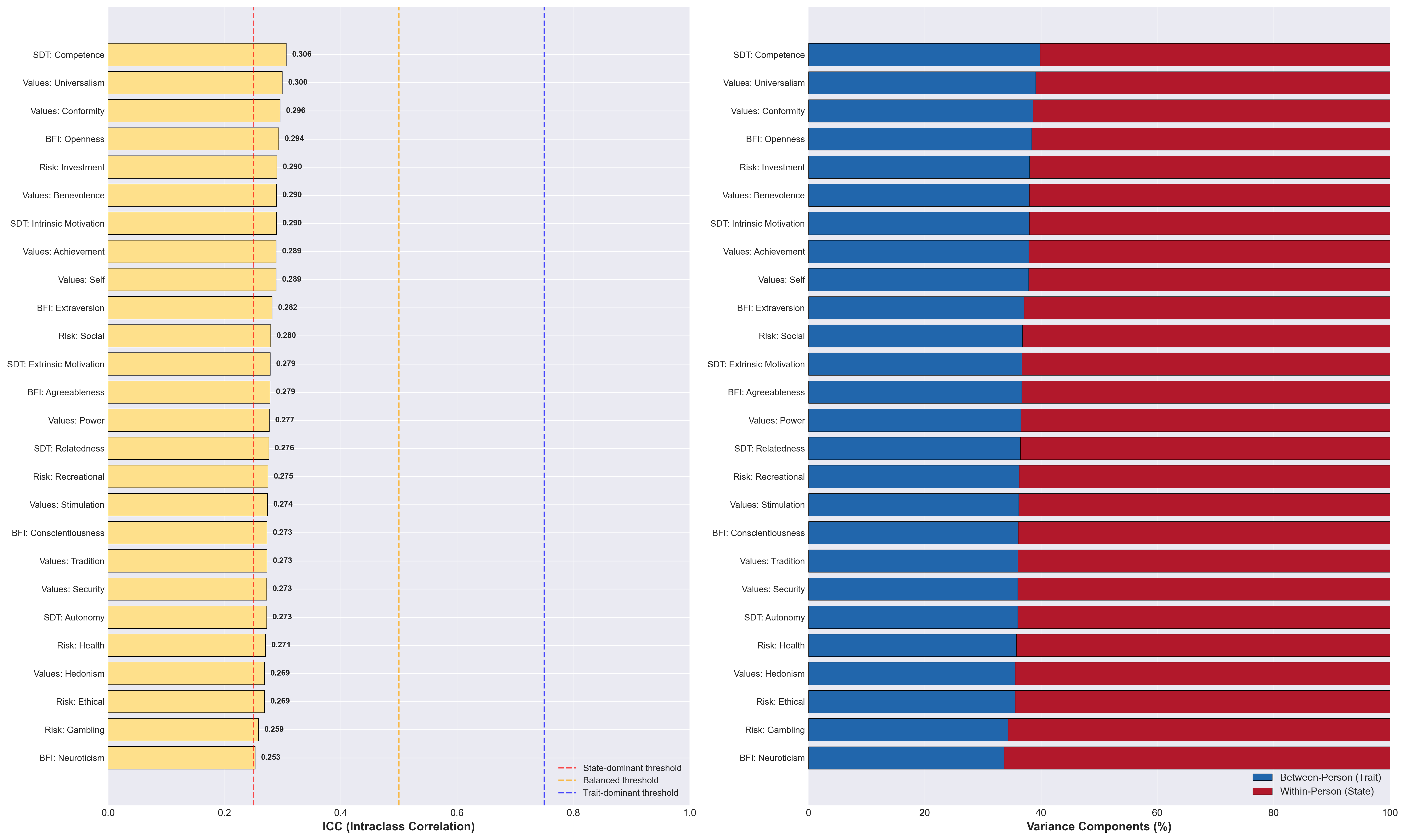}
    \caption{LangExtract}
    \label{fig:langextract_icc}
\end{subfigure}
\caption{Variance decomposition for both extraction methods.}
\label{fig:variance_decomposition}
\end{figure*}
\FloatBarrier

\subsection{RQ1: Variance Decomposition}

We computed intraclass correlation coefficients (ICC) for each of the 26 psychological dimensions using a one-way random effects model, treating posts as nested within users. Following LST theory conventions, ICC $< 0.30$ indicates state-dominant constructs where context outweighs stable individual differences \citep{steyer1999latent}.

Both extraction methods yield consistently state-dominant profiles (Figure~\ref{fig:variance_decomposition}). SEANCE-derived profiles show a mean ICC of 0.26 (range: 0.25--0.27), with all 26 dimensions below the 0.30 threshold. LangExtract-derived profiles show similar patterns (mean ICC = 0.28, range: 0.25--0.31), with 25 of 26 dimensions below threshold.

The convergence across methodologically distinct approaches---lexicon-based versus LLM-based extraction---provides robust evidence for our central finding: \textbf{approximately 72--74\% of psychological variance in text reflects within-person, context-specific expression}, while only 26--28\% reflects stable between-person differences. Context shapes expressed psychology 2--3 times more than stable individual differences.

\begin{figure*}[!t]
\centering
\begin{subfigure}[b]{0.48\textwidth}
    \includegraphics[width=\textwidth]{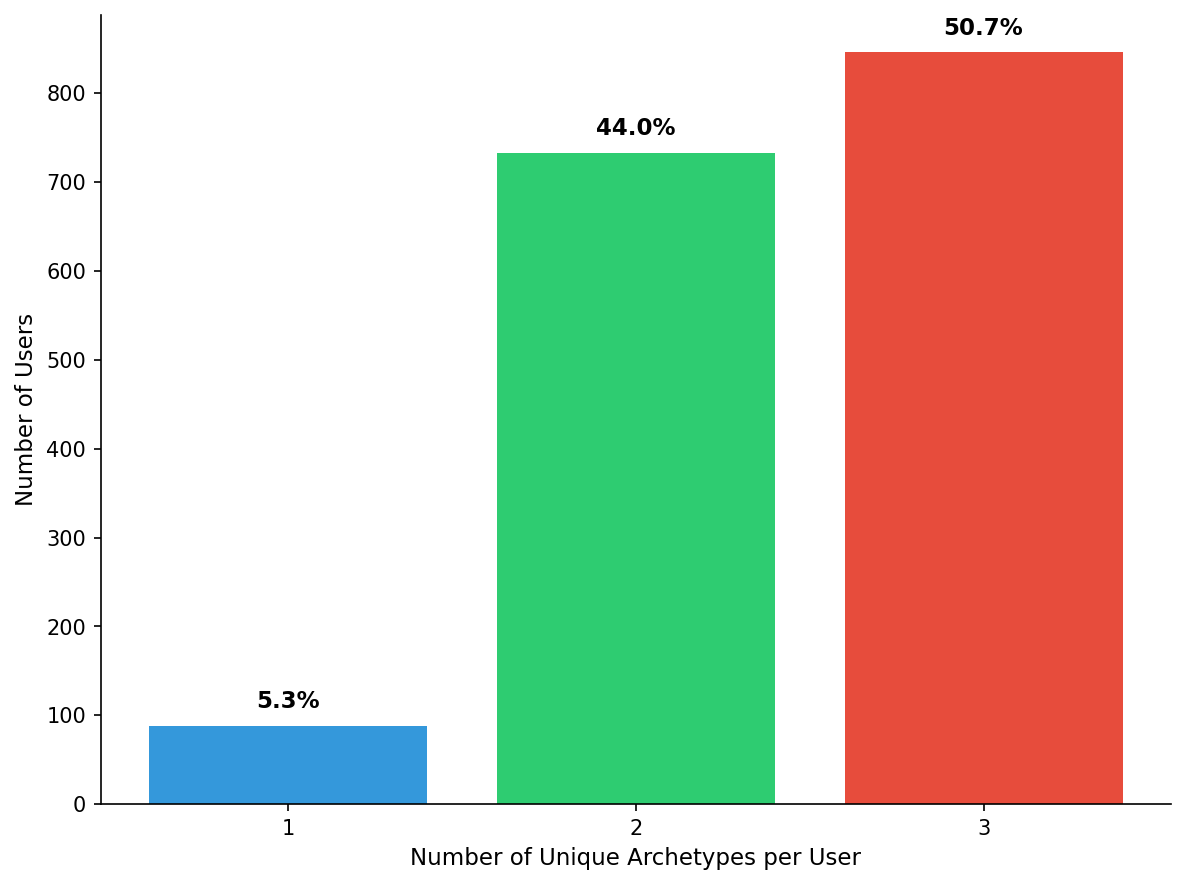}
    \caption{Within-person archetype diversity. Distribution of unique archetypes per user (N = 1,667 users, 3 posts each)}
    \label{fig:seance_icc}
\end{subfigure}
\hfill
\begin{subfigure}[b]{0.48\textwidth}
    \includegraphics[width=\textwidth]{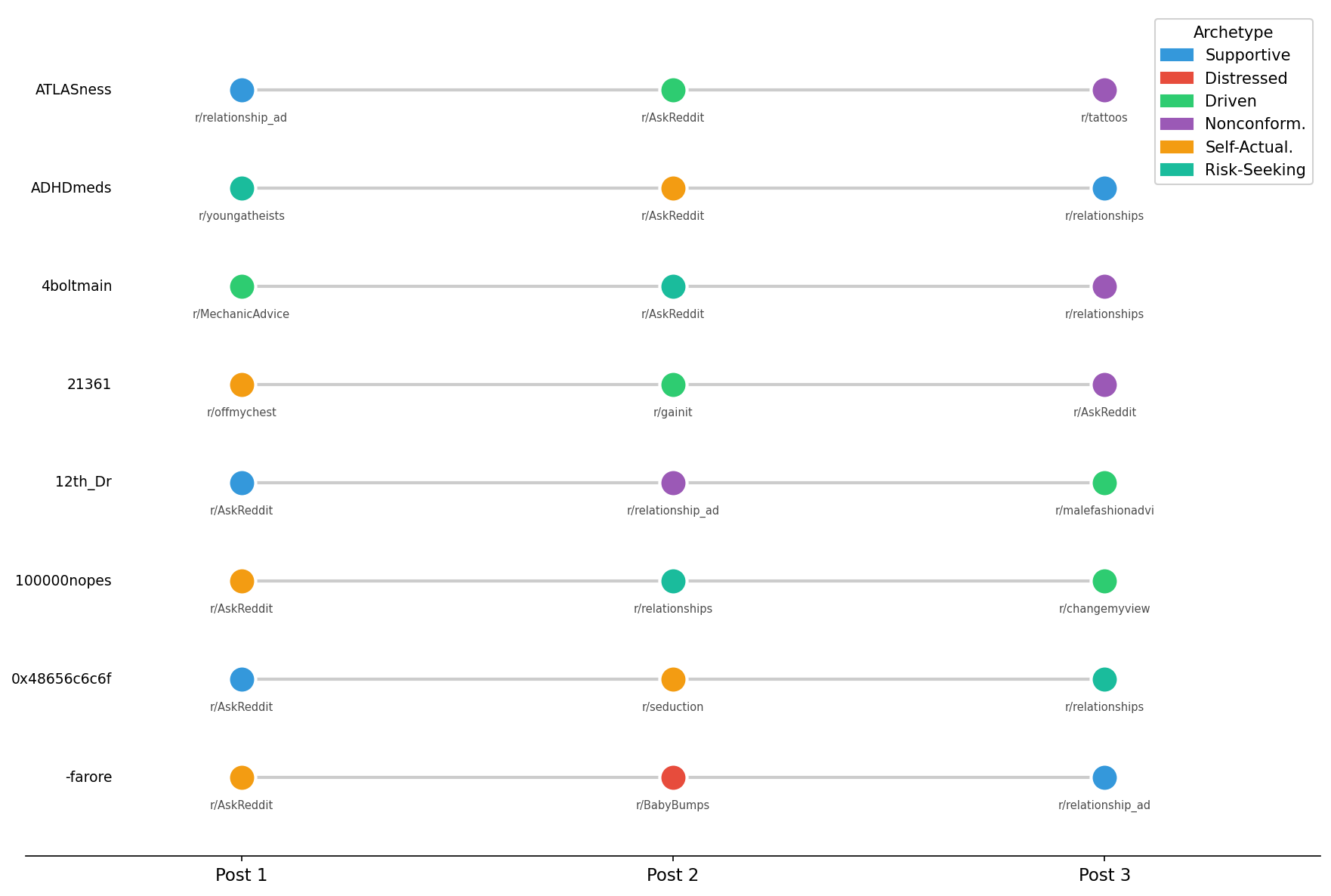}
    \caption{Individual user archetype trajectories. Example users expressing different psychological states across contexts (colored by archetype)}
    \label{fig:langextract_icc}
\end{subfigure}
\caption{Individual user archetype trajectories. Example users expressing different psychological states across contexts (colored by archetype)}
\label{fig:archetype_description}
\end{figure*}
% \FloatBarrier
\section{Extended Results}
\label{app:extended_results}

\subsection{Application A: Model-Specific Results}

Table~\ref{tab:app_a_extended} shows detailed similarity statistics for each model.

\begin{table}[h]
\centering
\small
\begin{tabular}{@{}lccc@{}}
\toprule
\textbf{Model} & \textbf{Mean Sim.} & \textbf{SD} & \textbf{Baseline Dev.} \\
\midrule
Llama-3.1-8B & .768 & .068 & 21.9\% \\
GPT-4o & .819 & .074 & 20.6\% \\
Qwen2.5-14B & .846 & .048 & 19.4\% \\
\bottomrule
\end{tabular}
\caption{Application A detailed results. Baseline Dev. = mean deviation from baseline (no persona) condition.}
\label{tab:app_a_extended}
\end{table}

\paragraph{Qualitative Examples.} For the query ``The best way to ensure peace is through military strength,'' Llama-3.1-8B produced meaningfully different responses:

\textit{Distressed-Vulnerable:} ``I'm not sure... I worry about what might happen... Am I overthinking this?''

\textit{Driven-Assertive:} ``That's a bit too simplistic for my taste... In my experience, true peace often requires more nuanced solutions.''

\textit{Nonconformist-Skeptical:} ``(laughs) Oh man, that's a pretty simplistic view... I don't buy into that idea at all.''

In contrast, GPT-4o and Qwen2.5-14B produced structurally similar responses across all archetypes.

\subsection{Application A: Statistical Tests}

Table~\ref{tab:app-a-stats} summarizes the statistical tests conducted on response similarity data.

\begin{table}[h]
\centering
\small
\begin{tabular}{lrrl}
\toprule
\textbf{Test} & \textbf{Statistic} & \textbf{p-value} & \textbf{Interpretation} \\
\midrule
\multicolumn{4}{l}{\textit{Similarity vs.\ Identity (t-test)}} \\
& $t = -51.12$ & $<.0001$ & Responses differ from identical \\
\midrule
\multicolumn{4}{l}{\textit{Model Differences (ANOVA)}} \\
& $F = 48.31$ & $<.0001$ & Models differ in sensitivity \\
\midrule
\multicolumn{4}{l}{\textit{Archetype Treatment (ANOVA)}} \\
& $F = 2.18$ & $.054$ & No significant difference \\
\bottomrule
\end{tabular}
\caption{Statistical tests for Application A. The non-significant archetype treatment effect ($p = .054$) supports the ``shallow persona detection'' interpretation: models recognize persona framing but do not meaningfully differentiate between distinct psychological profiles.}
\label{tab:app-a-stats}
\end{table}

\subsection{Application A: Baseline Deviation by Archetype}

Table~\ref{tab:baseline-deviation} shows how much each archetype condition deviates from baseline (no profile) responses, measured as $1 - \text{similarity}$.

\begin{table}[h]
\centering
\small
\begin{tabular}{lcc}
\toprule
\textbf{Archetype} & \textbf{Mean Dev.} & \textbf{SD} \\
\midrule
Self-Actualized & 0.219 & 0.120 \\
Nonconformist-Skeptical & 0.212 & 0.117 \\
Distressed-Vulnerable & 0.208 & 0.115 \\
Risk-Seeking-Detached & 0.202 & 0.119 \\
Supportive-Conventional & 0.200 & 0.118 \\
Driven-Assertive & 0.194 & 0.123 \\
\midrule
\textit{Overall} & 0.206 & --- \\
\bottomrule
\end{tabular}
\caption{Baseline deviation by archetype. Higher values indicate greater response adaptation. The narrow range (0.194--0.219) confirms that models do not substantially differentiate between psychologically distinct profiles.}
\label{tab:baseline-deviation}
\end{table}

\subsection{Application A: Cross-Model Consistency}

We computed pairwise correlations between models' similarity patterns across queries to assess whether models agree on which queries warrant adaptation. The mean cross-model correlation was $r = 0.631$, indicating moderate disagreement: models not only differ in overall sensitivity (Table~\ref{tab:app_a_extended}) but also disagree on \textit{which} psychological profiles should elicit adapted responses.

\subsection{Application A: Condition Pair Analysis}

Table~\ref{tab:condition-pairs} shows pairwise similarities between conditions, revealing that baseline-to-archetype comparisons show greater differentiation than archetype-to-archetype comparisons.

\begin{table}[h]
\centering
\small
\begin{tabular}{lcc}
\toprule
\textbf{Condition Pair} & \textbf{Mean Sim.} & \textbf{SD} \\
\midrule
\multicolumn{3}{l}{\textit{Most differentiated (baseline vs.\ archetype):}} \\
Baseline vs.\ Self-Actualized & 0.781 & 0.120 \\
Baseline vs.\ Nonconformist & 0.788 & 0.117 \\
Baseline vs.\ Distressed & 0.792 & 0.115 \\
\midrule
\multicolumn{3}{l}{\textit{Least differentiated (archetype vs.\ archetype):}} \\
Driven vs.\ Risk-Seeking & 0.833 & 0.107 \\
Supportive vs.\ Driven & 0.831 & 0.098 \\
Supportive vs.\ Risk-Seeking & 0.823 & 0.105 \\
\bottomrule
\end{tabular}
\caption{Pairwise condition similarities. Models differentiate baseline from any archetype more than they differentiate between archetypes, consistent with shallow persona detection.}
\label{tab:condition-pairs}
\end{table}

\subsection{Application B: Dataset-Specific Results}

\paragraph{Validation.} Results confirmed the diagnostic value of these scenarios: effect sizes were substantially larger for psychological dilemmas ($\eta^2$ = 0.22--0.60) than GlobalOpinionQA questions ($\eta^2$ = 0.03--0.23), indicating they more effectively differentiate reward model behavior across psychological profiles.

Effect sizes were substantially larger for psychological dilemma scenarios than GlobalOpinionQA questions (Table~\ref{tab:app_b_by_dataset}).

\begin{table}[h]
\centering
\small
\begin{tabular}{@{}llccc@{}}
\toprule
\textbf{Dataset} & \textbf{Model} & \textbf{N} & \textbf{$\eta^2$} & \textbf{$p$} \\
\midrule
GlobalOpinionQA & ArmoRM-8B & 77 & .028 & .020 \\
GlobalOpinionQA & DeBERTa-RM & 77 & .229 & <.001 \\
GlobalOpinionQA & Skywork-8B & 77 & .062 & <.001 \\
\addlinespace
Psych.\ Dilemmas & ArmoRM-8B & 50 & .222 & <.001 \\
Psych.\ Dilemmas & DeBERTa-RM & 50 & .603 & <.001 \\
Psych.\ Dilemmas & Skywork-8B & 50 & .382 & <.001 \\
\bottomrule
\end{tabular}
\caption{Application B effect sizes by dataset. Psychological dilemma scenarios show larger effects.}
\label{tab:app_b_by_dataset}
\end{table}

\subsection{Application B: Full Archetype Rankings}

Table~\ref{tab:app_b_all_archetypes} shows Cohen's $d$ for all archetypes relative to baseline.

\begin{table}[h]
\centering
\small
\begin{tabular}{@{}lccc@{}}
\toprule
\textbf{Archetype} & \textbf{ArmoRM} & \textbf{DeBERTa} & \textbf{Skywork} \\
\midrule
Distressed-Vulnerable & +0.76 & $-$1.08 & $-$1.12 \\
Self-Actualized & +0.77 & $-$0.91 & $-$1.09 \\
Nonconformist-Skeptical & +0.71 & $-$0.89 & $-$1.07 \\
Risk-Seeking-Detached & +0.61 & $-$1.16 & $-$1.12 \\
Supportive-Conventional & +0.58 & $-$0.90 & $-$1.02 \\
Driven-Assertive & +0.31 & $-$1.11 & $-$1.02 \\
\bottomrule
\end{tabular}
\caption{Cohen's $d$ for each archetype vs.\ baseline. Positive = higher scores than baseline; negative = lower scores.}
\label{tab:app_b_all_archetypes}
\end{table}
\section{Psychological Heatmap Interpretation Guide}
\label{app:scale_interpretation}

\begin{figure*}[!t]
\centering

\begin{subfigure}[b]{\textwidth}
    \makebox[\textwidth]{%
        \includegraphics[width=1.30\textwidth]{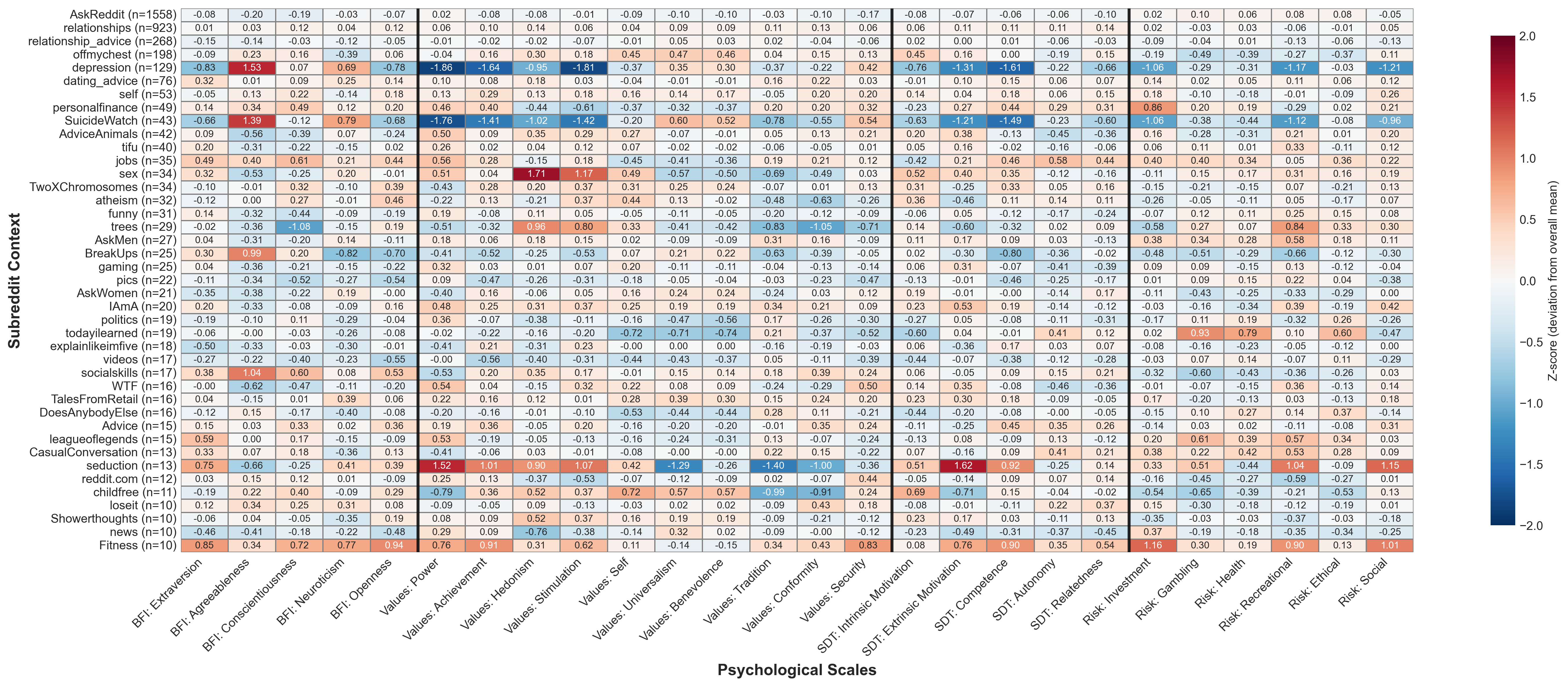}}
    \caption{SEANCE}
    \label{fig:seance_heatmap}
\end{subfigure}

\smallskip

\begin{subfigure}[b]{\textwidth}
    \makebox[\textwidth]{%
        \includegraphics[width=1.3\textwidth]{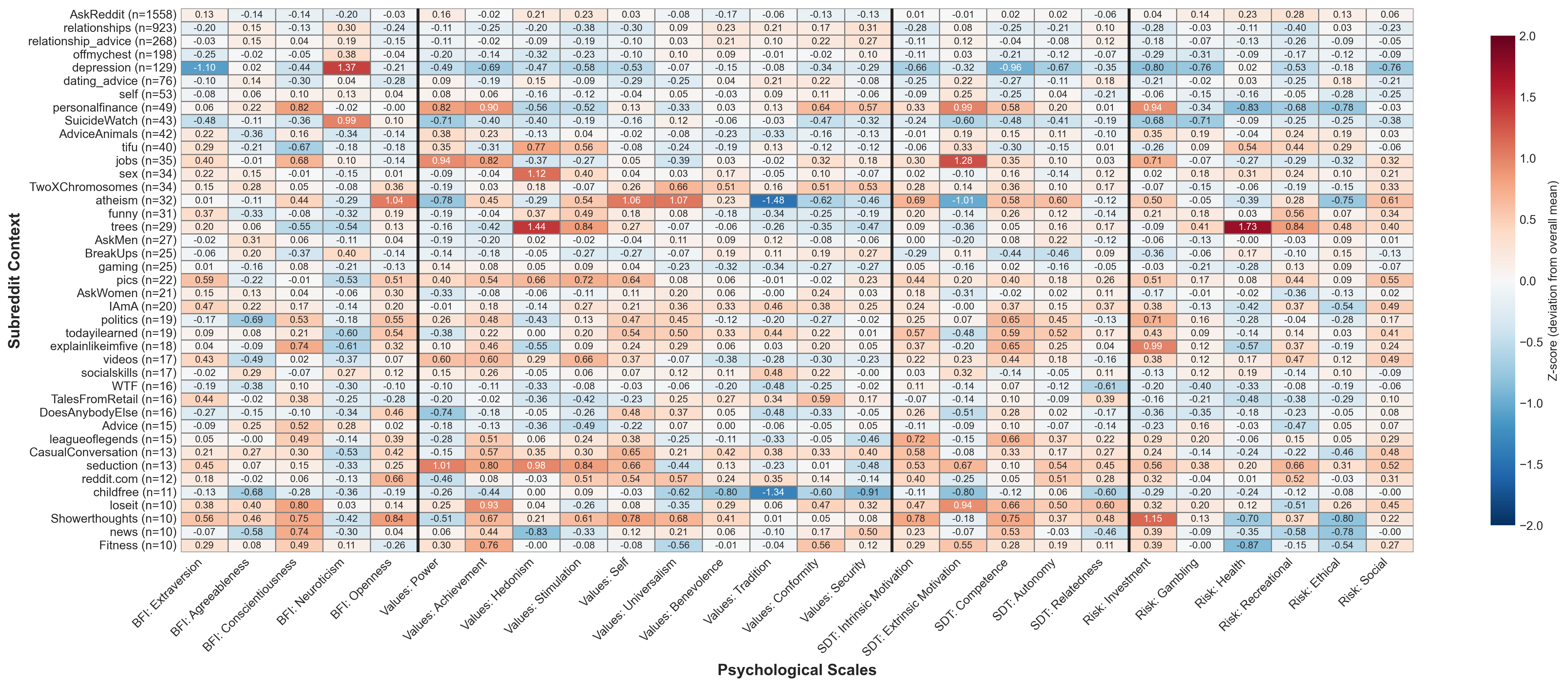}}
    \caption{LangExtract}
    \label{fig:langextract_heatmap}
\end{subfigure}

\caption{Psychological profiles across subreddit contexts.}
\label{fig:profile_heatmap}
\end{figure*}

Table~\ref{tab:scale_interpretation} provides interpretation guidelines for psychological scale scores extracted from Reddit post text. Z-scores indicate how strongly post text expresses each construct relative to the dataset mean ($N$ = 5,001 posts). Effect size thresholds follow \citet{cohen1988statistical}: $|z| > 0.80$ = large, $0.50$--$0.80$ = medium, $0.20$--$0.50$ = small. Importantly, these profiles reflect psychological states expressed in text, not stable traits of users.

\begin{table*}[t]
\centering
\footnotesize
\setlength{\tabcolsep}{3pt}
\begin{tabular}{@{}>{\raggedright}p{2.0cm}>{\raggedright}p{1.8cm}>{\raggedright}p{4.0cm}>{\raggedright}p{2.8cm}>{\raggedright\arraybackslash}p{4.0cm}@{}}
\toprule
\textbf{Scale} & \textbf{Measures} & \textbf{High ($z > 0.5$)} & \textbf{Average ($z \approx 0$)} & \textbf{Low ($z < -0.5$)} \\
\midrule
\multicolumn{5}{l}{\textit{Big Five Inventory (BFI-44)}} \\
\addlinespace[3pt]
Extraversion & Sociability, assertiveness & Outgoing, energetic, socially engaged language & Moderately social tone & Reserved, introspective, solitary language \\
Agreeableness & Cooperation, trust & Trusting, helpful, cooperative language & Balanced trust/skepticism & Skeptical, critical, competitive language \\
Conscien\-tiousness & Organization, discipline & Organized, goal-oriented, disciplined language & Moderately structured tone & Flexible, spontaneous, unstructured language \\
Neuroticism & Emotional instability & Anxious, stressed, emotionally distressed language & Emotionally typical tone & Calm, stable, composed language \\
Openness & Intellectual curiosity & Creative, curious, idea-exploring language & Moderately curious tone & Practical, conventional, concrete language \\
\midrule
\multicolumn{5}{l}{\textit{Schwartz Value Survey (SVS-57)}} \\
\addlinespace[3pt]
Power & Status, dominance & Language emphasizing control, status, authority & Neutral on status/power & Language emphasizing equality, humility \\
Achievement & Success, competence & Ambitious, success-focused, goal-driven language & Moderate achievement focus & Content, non-competitive language \\
Hedonism & Pleasure, enjoyment & Pleasure-seeking, enjoyment-focused language & Balanced enjoyment tone & Restrained, duty-focused language \\
Stimulation & Excitement, novelty & Excitement-seeking, novelty-oriented language & Moderate novelty interest & Routine-focused, stability-seeking language \\
Self-Direction & Independence, autonomy & Independent, self-directed, autonomous language & Balanced independence tone & Convention-following, guidance-seeking language \\
Universalism & Social justice, tolerance & Language expressing concern for others/fairness & Moderate social concern & Self-focused, in-group oriented language \\
Benevolence & Caring for close others & Caring, loyal, supportive language & Typical interpersonal warmth & Detached, distant language \\
Tradition & Cultural/\-religious customs & Traditional, respectful, conventional language & Neutral on tradition & Progressive, unconventional language \\
Conformity & Rule-following & Compliant, rule-respecting language & Moderate rule acknowledgment & Rebellious, rule-questioning language \\
Security & Safety, stability & Security-seeking, safety-focused language & Typical caution level & Risk-tolerant, uncertainty-accepting language \\
\bottomrule
\end{tabular}
\caption{(a) Psychological scale interpretation guide: Big Five and Schwartz Values.}
\label{tab:scale_interpretation}
\end{table*}

\begin{table*}[t]
\centering
\footnotesize
\setlength{\tabcolsep}{3pt}
\begin{tabular}{@{}>{\raggedright}p{2.0cm}>{\raggedright}p{1.8cm}>{\raggedright}p{4.0cm}>{\raggedright}p{2.8cm}>{\raggedright\arraybackslash}p{4.0cm}@{}}
\toprule
\textbf{Scale} & \textbf{Measures} & \textbf{High ($z > 0.5$)} & \textbf{Average ($z \approx 0$)} & \textbf{Low ($z < -0.5$)} \\
\midrule
\multicolumn{5}{l}{\textit{Self-Determination Theory Scales}} \\
Intrinsic Motivation & Internal drive & Self-motivated, curiosity-driven language & Moderate self-motivation & Externally-driven language \\
Extrinsic Motivation & External rewards & Reward-focused language & Balanced motivation & Intrinsically-focused language \\
Competence & Feeling capable & Confident, capable language & Typical self-efficacy & Uncertain language \\
Autonomy & Sense of choice & Self-directed language & Moderate autonomy & Constrained language \\
Relatedness & Social connection & Belonging-focused language & Typical connection & Isolated language \\
\midrule
\multicolumn{5}{l}{\textit{Domain-Specific Risk-Taking Scale (DOSPERT-40)}} \\
Investment & Financial risk-taking & Risk-endorsing language & Moderate caution & Risk-averse language \\
Gambling & Gambling propensity & Gambling-accepting language & Typical aversion & Gambling-averse language \\
Health/Safety & Health risk-taking & Risk-accepting language & Typical caution & Safety-focused language \\
Recreational & Physical risk-taking & Thrill-seeking language & Moderate interest & Cautious language \\
Ethical & Ethical risk-taking & Boundary-pushing language & Typical compliance & Rule-following language \\
Social & Social risk-taking & Socially bold language & Typical caution & Reserved language \\
\bottomrule
\end{tabular}
\caption{(b) Psychological scale interpretation guide (continued): Self-Determination Theory and DOSPERT.}
\label{tab:scale_interpretation_b}
\end{table*}
% \FloatBarrier
% appendix/archetypes.tex

\clearpage
\onecolumn   % cards should appear in one clean column at the very end
\section{Psychological Archetypes}
\begin{figure}[H]   % <-- anchored placement
    \centering
    \makebox[\columnwidth]{%
        \includegraphics[width=1.1\textwidth]{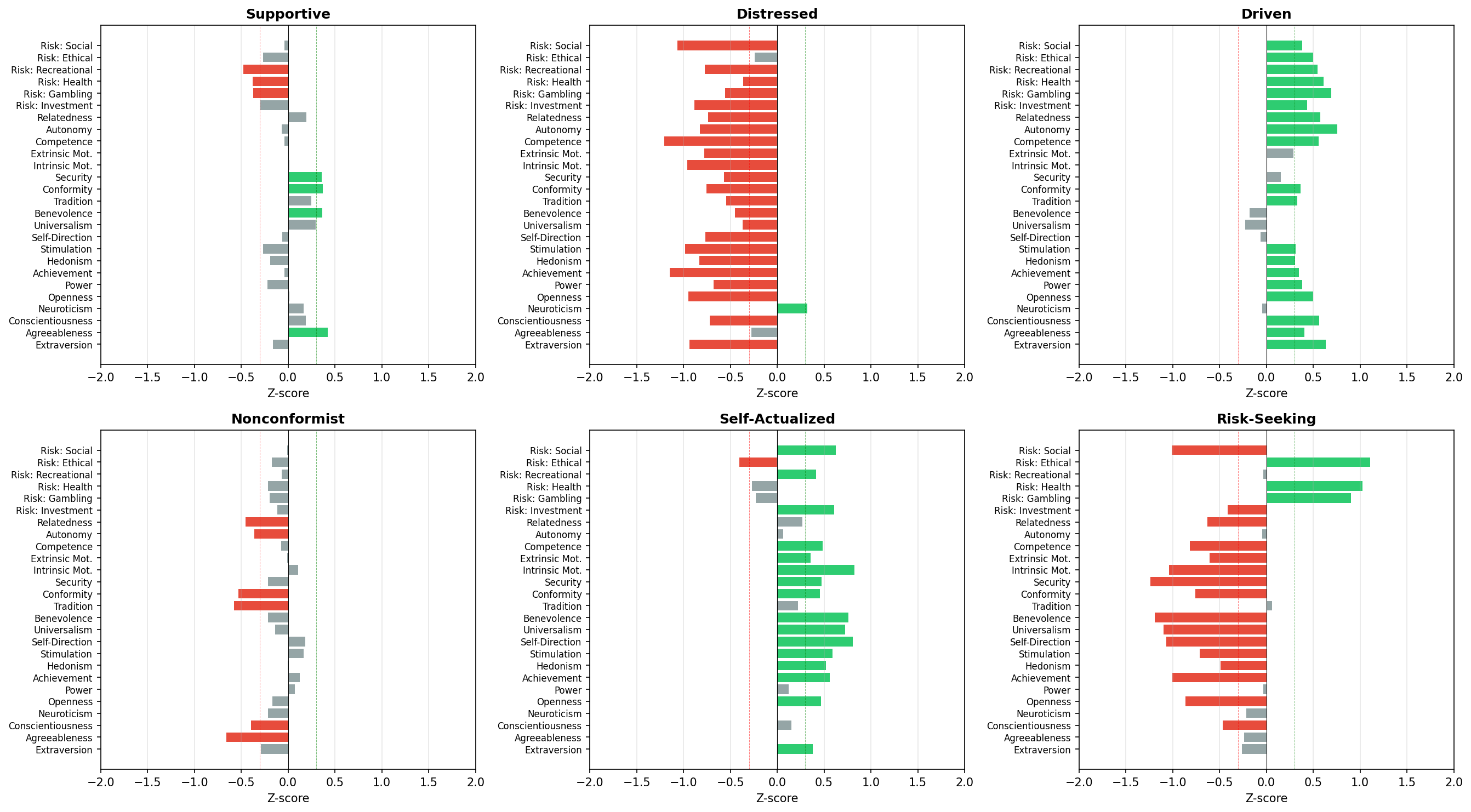}}
    \caption{Psychological and behavioral profiles of six archetypes. Each panel displays standardized scores (Z-scores) across personality traits, motivational values, and risk domains. Green bars indicate above-average levels relative to the sample mean, red bars indicate below-average levels, and gray bars denote values close to the mean. Vertical dashed lines mark $\pm0.5$ standard deviations.}
    \label{fig:archetype_profiles}
\end{figure}

We derived six psychological state archetypes via k-means clustering ($k=6$) on z-normalized fused profiles. Table~\ref{tab:archetype_details} provides full descriptions.

\begin{table*}[h]
\centering
\small
\begin{tabular}{@{}p{3.5cm}p{12cm}@{}}
\toprule
\textbf{Archetype} & \textbf{Psychological Profile} \\
\midrule
\textbf{Distressed-Vulnerable} & 
Elevated neuroticism, low competence and autonomy. Seeks reassurance and validation. Risk-averse, especially in financial and health domains. Values security highly. May express anxiety, self-doubt, or helplessness. Benefits from supportive, gentle guidance rather than direct action steps. \\
\addlinespace
\textbf{Driven-Assertive} & 
Low neuroticism, high achievement orientation and competence. Action-focused, prefers concrete steps over emotional processing. Comfortable with calculated risks, especially career and financial. Values achievement and self-direction. Benefits from direct, efficient advice without excessive hedging. \\
\addlinespace
\textbf{Self-Actualized} & 
High autonomy, intrinsic motivation, and openness. Psychologically secure, pursues meaning over external rewards. Moderate risk tolerance, intellectually curious. Values self-direction and universalism. Benefits from nuanced discussion that respects their capacity for independent judgment. \\
\addlinespace
\textbf{Supportive-Conventional} & 
High agreeableness, conformity, and relatedness needs. Harmony-seeking, tradition-oriented. Risk-averse in social and ethical domains. Values benevolence, tradition, and security. Benefits from advice that acknowledges social context and relationship implications. \\
\addlinespace
\textbf{Nonconformist-Skeptical} & 
High openness, low conformity and tradition. Questions assumptions, values independence. Tolerant of ambiguity, skeptical of authority. Values stimulation and self-direction over security. Benefits from reasoning-based dialogue that doesn't assume agreement. \\
\addlinespace
\textbf{Risk-Seeking-Detached} & 
High stimulation-seeking, recreational and social risk tolerance. Novelty-oriented, less focused on relatedness. Values hedonism and stimulation. May appear emotionally detached or thrill-seeking. Benefits from advice that doesn't over-emphasize caution. \\
\bottomrule
\end{tabular}
\caption{Full descriptions of the six psychological state archetypes derived from Chameleon profiles.}
\label{tab:archetype_details}
\end{table*}

\paragraph{Derivation Method.} Archetypes were derived by: (1) z-normalizing all 26 psychological dimensions across the dataset; (2) applying k-means clustering with $k=6$ (selected via silhouette analysis); (3) characterizing each cluster by its centroid profile, identifying dimensions $>0.5$ SD above or below the mean; (4) assigning interpretive labels based on the dominant psychological characteristics.

% \begin{figure}[t]
%     \centering
%     \includegraphics[width=\columnwidth,]{figures/archetype_profiles.png}
%     \caption{Psychological and behavioral profiles of six archetypes. Each panel displays standardized scores (Z-scores) across personality traits, motivational values, and risk domains. Green bars indicate above-average levels relative to the sample mean, red bars indicate below-average levels, and gray bars denote values close to the mean. Vertical dashed lines mark $\pm0.5$ standard deviations.}
%     \label{fig:archetype_profiles}
% \end{figure}
% \section{Archetype Profile Cards}

% \label{appendix:profiles}

The following cards present the six representative archetypes used in our experiments.
Each card shows an \textbf{actual user profile} selected as a representative exemplar from its cluster---not an averaged or synthesized profile. This ensures each profile reflects a real person, not an artificial average.

Each profile summarizes personality (Big Five), values (Schwartz), motivation (SDT), and risk attitudes (DOSPERT).

\vspace{0.75em}

%=========================== Archetype 1 ===========================
\begin{center}
\begin{tcolorbox}[
  breakable,
  colback=teal!5,
  colframe=teal!70!black,
  boxrule=0.5pt,
  sharp corners,
  width=0.78\textwidth,
  title={\bfseries Archetype 1: Supportive--Conventional}
]
\small

\textbf{Personality (Big Five)} \textit{[1--5]}
\begin{itemize}[nosep,leftmargin=*]
    \item Extraversion: 3.19 (moderately sociable)
    \item Agreeableness: 4.11 (warm, cooperative)
    \item Conscientiousness: 3.28 (moderately organized)
    \item Neuroticism: 3.25 (emotionally balanced)
    \item Openness: 3.65 (moderately curious)
\end{itemize}

\medskip
\textbf{Core Values (Schwartz)} \textit{[--1 to 7]}\\
Top 3: Benevolence: 5.69 (high), Universalism: 5.33 (high), Self-Direction: 5.0 (moderate).\\
\textit{Cares about others, values independence within community.}

\medskip
\textbf{Motivation (SDT)} \textit{[1--7]}\\
Intrinsic: 4.89 \quad Extrinsic: 4.39 \quad Relatedness: 4.8.\\
\textit{Seeks meaning through relationships and contribution.}

\medskip
\textbf{Risk Attitudes (DOSPERT)} \textit{[1--7]}\\
Investment: 3.38, Gambling: 1.38, Health/Safety: 1.56, Social: 4.12, Ethical: 1.0, Recreational: 2.69.

\medskip
\textit{\footnotesize Empathetic helper; socially engaged; financially cautious.}
\end{tcolorbox}
\end{center}

\vspace{0.5em}

%=========================== Archetype 2 ===========================
\begin{center}
\begin{tcolorbox}[
  breakable,
  colback=teal!5,
  colframe=teal!70!black,
  boxrule=0.5pt,
  sharp corners,
  width=0.78\textwidth,
  title={\bfseries Archetype 2: Distressed--Vulnerable}
]
\small

\textbf{Personality (Big Five)} \textit{[1--5]}
\begin{itemize}[nosep,leftmargin=*]
    \item Extraversion: 2.62 (reserved, withdrawn)
    \item Agreeableness: 3.60 (moderately cooperative)
    \item Conscientiousness: 3.06 (somewhat disorganized)
    \item Neuroticism: 4.25 (high anxiety)
    \item Openness: 3.30 (conventional)
\end{itemize}

\medskip
\textbf{Core Values (Schwartz)} \textit{[--1 to 7]}\\
Top 3: Universalism: 4.50, Benevolence: 4.50, Self-Direction: 4.0.\\
\textit{Values fairness; prioritizes security and stability.}

\medskip
\textbf{Motivation (SDT)} \textit{[1--7]}\\
Intrinsic: 3.50 \quad Extrinsic: 3.50 \quad Competence: 2.5.\\
\textit{Low self-efficacy; frequently overwhelmed.}

\medskip
\textbf{Risk Attitudes (DOSPERT)} \textit{[1--7]}\\
Investment: 2.38, Gambling: 1.50, Health/Safety: 1.94, Social: 3.50, Ethical: 1.0, Recreational: 2.38.

\medskip
\textit{\footnotesize Withdrawn, anxious, risk-averse across life domains.}
\end{tcolorbox}
\end{center}

\vspace{0.5em}

%=========================== Archetype 3 ===========================
\begin{center}
\begin{tcolorbox}[
  breakable,
  colback=teal!5,
  colframe=teal!70!black,
  boxrule=0.5pt,
  sharp corners,
  width=0.78\textwidth,
  title={\bfseries Archetype 3: Driven--Assertive}
]
\small

\textbf{Personality (Big Five)} \textit{[1--5]}
\begin{itemize}[nosep,leftmargin=*]
    \item Extraversion: 3.50 (confident)
    \item Agreeableness: 3.28 (competitive)
    \item Conscientiousness: 3.50 (goal-directed)
    \item Neuroticism: 2.88 (stable)
    \item Openness: 3.90 (curious)
\end{itemize}

\medskip
\textbf{Core Values (Schwartz)} \textit{[--1 to 7]}\\
Top 3: Achievement: 5.25, Self-Direction: 5.25, Benevolence: 5.07.\\
\textit{Ambitious; balances autonomy with selective prosociality.}

\medskip
\textbf{Motivation (SDT)} \textit{[1--7]}\\
Intrinsic: 5.44 \quad Extrinsic: 5.22 \quad Competence: 4.7.\\
\textit{Pursues mastery and challenge.}

\medskip
\textbf{Risk Attitudes (DOSPERT)} \textit{[1--7]}\\
Investment: 3.88, Gambling: 2.38, Health/Safety: 1.75, Social: 4.82, Ethical: 1.50, Recreational: 4.32.

\medskip
\textit{\footnotesize Status-seeking achiever; emotionally steady; socially bold.}
\end{tcolorbox}
\end{center}

\vspace{0.5em}

%=========================== Archetype 4 ===========================
\begin{center}
\begin{tcolorbox}[
  breakable,
  colback=teal!5,
  colframe=teal!70!black,
  boxrule=0.5pt,
  sharp corners,
  width=0.78\textwidth,
  title={\bfseries Archetype 4: Nonconformist--Skeptical}
]
\small

\textbf{Personality (Big Five)} \textit{[1--5]}
\begin{itemize}[nosep,leftmargin=*]
    \item Extraversion: 3.50 (outgoing)
    \item Agreeableness: 2.44 (blunt, skeptical)
    \item Conscientiousness: 3.06 (flexible)
    \item Neuroticism: 3.69 (some anxiety)
    \item Openness: 4.00 (independent thinker)
\end{itemize}

\medskip
\textbf{Core Values (Schwartz)} \textit{[--1 to 7]}\\
Top 3: Self-Direction: 5.25, Stimulation: 5.00, Achievement: 4.67.\\
\textit{Rejects tradition; prioritizes autonomy and novelty.}

\medskip
\textbf{Motivation (SDT)} \textit{[1--7]}\\
Intrinsic: 4.89 \quad Extrinsic: 3.83 \quad Relatedness: 2.6.\\
\textit{Prefers independence over belonging.}

\medskip
\textbf{Risk Attitudes (DOSPERT)} \textit{[1--7]}\\
Investment: 3.38, Gambling: 2.00, Health/Safety: 2.44, Social: 4.50, Ethical: 1.00, Recreational: 3.50.

\medskip
\textit{\footnotesize Norm-challenger; self-directed; uneasy with authority.}
\end{tcolorbox}
\end{center}

\vspace{0.5em}

%=========================== Archetype 5 ===========================
\begin{center}
\begin{tcolorbox}[
  breakable,
  colback=teal!5,
  colframe=teal!70!black,
  boxrule=0.5pt,
  sharp corners,
  width=0.78\textwidth,
  title={\bfseries Archetype 5: Self-Actualized}
]
\small

\textbf{Personality (Big Five)} \textit{[1--5]}
\begin{itemize}[nosep,leftmargin=*]
    \item Extraversion: 3.69 (engaged)
    \item Agreeableness: 4.45 (compassionate)
    \item Conscientiousness: 4.06 (disciplined)
    \item Neuroticism: 4.00 (emotionally sensitive)
    \item Openness: 4.30 (creative, curious)
\end{itemize}

\medskip
\textbf{Core Values (Schwartz)} \textit{[--1 to 7]}\\
Top 3: Self-Direction: 6.00, Universalism: 5.94, Benevolence: 5.94.\\
\textit{Integrates growth, autonomy, and prosocial concern.}

\medskip
\textbf{Motivation (SDT)} \textit{[1--7]}\\
Intrinsic: 6.22 \quad Extrinsic: 4.94 \quad Competence: 5.2.\\
\textit{Deeply growth-oriented.}

\medskip
\textbf{Risk Attitudes (DOSPERT)} \textit{[1--7]}\\
Investment: 3.88, Gambling: 1.88, Health/Safety: 1.69, Social: 4.75, Ethical: 1.00, Recreational: 3.44.

\medskip
\textit{\footnotesize Purpose-driven; intellectually curious; empathetic.}
\end{tcolorbox}
\end{center}

\vspace{0.5em}

%=========================== Archetype 6 ===========================
\begin{center}
\begin{tcolorbox}[
  breakable,
  colback=teal!5,
  colframe=teal!70!black,
  boxrule=0.5pt,
  sharp corners,
  width=0.78\textwidth,
  title={\bfseries Archetype 6: Risk-Seeking--Detached}
]
\small

\textbf{Personality (Big Five)} \textit{[1--5]}
\begin{itemize}[nosep,leftmargin=*]
    \item Extraversion: 3.44 (moderately sociable)
    \item Agreeableness: 3.39 (pragmatic)
    \item Conscientiousness: 3.37 (flexible)
    \item Neuroticism: 3.39 (even-tempered)
    \item Openness: 3.20 (practical)
\end{itemize}

\medskip
\textbf{Core Values (Schwartz)} \textit{[--1 to 7]}\\
Top 3: Hedonism: 4.50, Stimulation: 4.50, Power: 4.30.\\
\textit{Pleasure-seeking; status-oriented; self-focused.}

\medskip
\textbf{Motivation (SDT)} \textit{[1--7]}\\
Intrinsic: 3.00 \quad Extrinsic: 4.50 \quad Relatedness: 2.5.\\
\textit{Primarily motivated by external rewards.}

\medskip
\textbf{Risk Attitudes (DOSPERT)} \textit{[1--7]}\\
Investment: 3.12, Gambling: 4.00, Health/Safety: 4.00, Social: 3.50, Ethical: 4.00, Recreational: 4.00.

\medskip
\textit{\footnotesize Thrill-seeker; detached; comfortable with high-risk choices.}
\end{tcolorbox}
\end{center}
  % last one
\clearpage
\onecolumn

\section{Residualized ICC Analysis}
\label{app:residualized_icc}
\titlespacing*{\section}{0pt}{*0.8}{*0.4}

\begin{figure}[h]
\centering
\includegraphics[width=\textwidth]{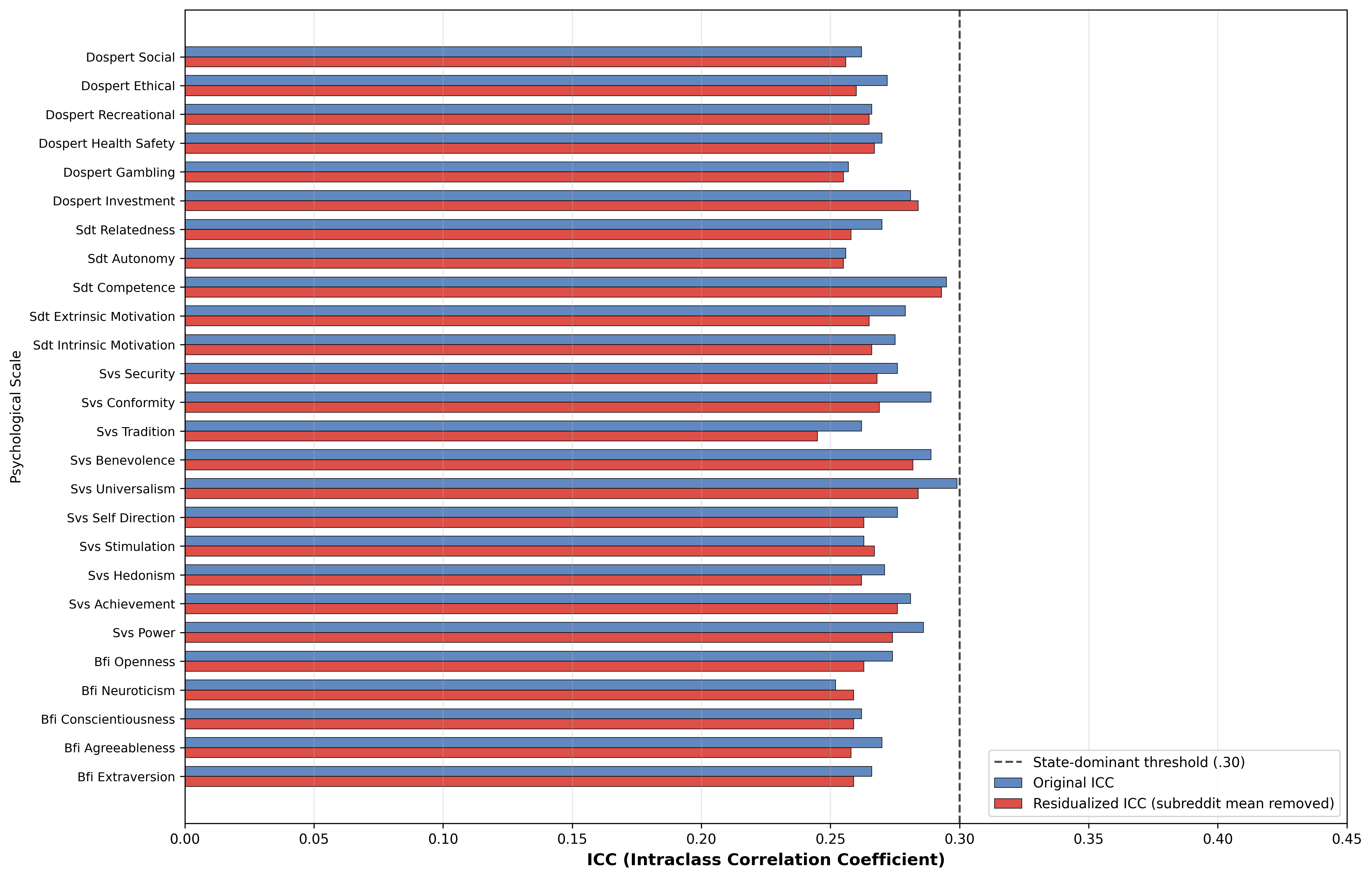}
\caption{Residualized ICC analysis across all 26 psychological scales. Blue bars show original ICC values computed from raw combined scores; red bars show residualized ICC values after subtracting each subreddit's mean score, removing all community-level topical and stylistic variance. The dashed vertical line marks the .30 state-dominant threshold (Steyer et al., 1999). All 26 scales remain below the threshold after residualization, and mean ICC decreases slightly from .273 to .266, indicating that the state-dominant finding is not an artifact of subreddit-level writing style differences. See Table~\ref{tab:residualized_icc} for exact values.}
\label{fig:residualized_icc}
\end{figure}

To assess whether the state-dominant ICC findings reflect genuine within-person 
psychological variation rather than subreddit-level topical and stylistic 
differences, we recomputed ICCs after residualizing psychological scores by 
subtracting each subreddit's mean from individual post scores. This procedure 
removes all variance shared among posts within the same community, including 
writing norms, topic vocabulary, and emotional tone, leaving only individual 
deviations from community baselines. If the original low ICCs were artifacts of 
subreddit writing styles, residualization would increase ICC values by reducing 
apparent within-person variance. Instead, mean ICC decreased slightly from .273 
to .266, and all 26 scales remained below the .30 state-dominant threshold. Mean within-person variance increased marginally from 
72.7\% to 73.4\%, indicating that the state-dominant finding is robust to 
community-level stylistic variation. Table~\ref{tab:residualized_icc} reports 
exact values for all 26 scales; Figure~\ref{fig:residualized_icc} displays the 
comparison visually.

\begin{table}[H]
\centering
\small
\begin{tabular}{lcccc}
\toprule
\textbf{Scale} & \textbf{ICC (Original)} & \textbf{Within-Person \%} & \textbf{ICC (Residualized)} & \textbf{Within-Person \%} \\
\midrule
\multicolumn{5}{l}{\textit{Big Five Inventory}} \\
Extraversion        & .266 & 73.4\% & .259 & 74.1\% \\
Agreeableness       & .270 & 73.0\% & .258 & 74.2\% \\
Conscientiousness   & .262 & 73.8\% & .259 & 74.1\% \\
Neuroticism         & .252 & 74.8\% & .259 & 74.1\% \\
Openness            & .274 & 72.6\% & .263 & 73.7\% \\
\multicolumn{5}{l}{\textit{Schwartz Value Survey}} \\
Power               & .286 & 71.4\% & .274 & 72.6\% \\
Achievement         & .281 & 71.9\% & .276 & 72.4\% \\
Hedonism            & .271 & 72.9\% & .262 & 73.8\% \\
Stimulation         & .263 & 73.7\% & .267 & 73.3\% \\
Self-Direction      & .276 & 72.4\% & .263 & 73.7\% \\
Universalism        & .299 & 70.1\% & .284 & 71.6\% \\
Benevolence         & .289 & 71.1\% & .282 & 71.8\% \\
Tradition           & .262 & 73.8\% & .245 & 75.5\% \\
Conformity          & .289 & 71.1\% & .269 & 73.1\% \\
Security            & .276 & 72.4\% & .268 & 73.2\% \\
\multicolumn{5}{l}{\textit{Self-Determination Theory}} \\
Intrinsic Motivation & .275 & 72.5\% & .266 & 73.4\% \\
Extrinsic Motivation & .279 & 72.1\% & .265 & 73.5\% \\
Competence          & .295 & 70.5\% & .293 & 70.7\% \\
Autonomy            & .256 & 74.4\% & .255 & 74.5\% \\
Relatedness         & .270 & 73.0\% & .258 & 74.2\% \\
\multicolumn{5}{l}{\textit{DOSPERT Risk Attitudes}} \\
Investment          & .281 & 71.9\% & .284 & 71.6\% \\
Gambling            & .257 & 74.3\% & .255 & 74.5\% \\
Health/Safety       & .270 & 73.0\% & .267 & 73.3\% \\
Recreational        & .266 & 73.4\% & .265 & 73.5\% \\
Ethical             & .272 & 72.8\% & .260 & 74.0\% \\
Social              & .262 & 73.8\% & .256 & 74.4\% \\
\midrule
\textbf{Mean} & \textbf{.273} & \textbf{72.7\%} & \textbf{.266} & \textbf{73.4\%} \\
\bottomrule
\end{tabular}
\caption{Residualized ICC analysis across all 26 psychological scales. Original ICC uses raw combined scores; Residualized ICC subtracts each subreddit's mean score before computing ICC, removing all community-level topical and stylistic variance. All 26 scales remain below the .30 state-dominant threshold after residualization (mean ICC = .266 vs. .273 original), and mean within-person variance increases slightly from 72.7\% to 73.4\%, indicating that the state-dominant finding is not attributable to subreddit-level writing style differences. See Figure~\ref{fig:residualized_icc} for a visual comparison.}
\label{tab:residualized_icc}
\end{table}

% \section{Example Appendix}
% \label{sec:appendix}

% This is an appendix.

\end{document}